\documentclass[manuscript,screen,nonacm]{acmart}
\renewcommand\footnotetextcopyrightpermission[1]{} 
\usepackage{enumitem}
\usepackage{placeins}
\usepackage{subcaption}
\usepackage[switch]{lineno}
\usepackage{scalerel}
\usepackage{multirow}
\usepackage{listings}
\usepackage{xcolor}
\usepackage{amsthm}
\usepackage{amsmath}
\usepackage{booktabs}
\usepackage{seqsplit}
\newtheorem{definition}{Definition}
\makeatletter
\def\@bibitem#1{\item\parskip 0pt\relax}
\makeatother

\ifodd 1

\newcommand{\com}[1]{\textbf{\color{red}(comment: #1)}} 
\newcommand{\res}[1]{\textbf{\color{magenta}(RESPONSE: #1)}} 
\else

\newcommand{\com}[1]{}
\newcommand{\res}[1]{}
\fi
\AtBeginDocument{%
  }

\setcopyright{acmlicensed}
\copyrightyear{2025}
\acmYear{2025}
\acmDOI{XXXXXXX.XXXXXXX}
\begin{document}

\title{How Strategic Agents Respond: Comparing Analytical Models with LLM-Generated Responses in Strategic Classification}

\author{Tian Xie}
\authornote{Both authors contributed equally to this research.}
\email{xie.1379@osu.edu}
\author{Pavan Rauch}
\authornotemark[1]
\email{rauch.139@osu.edu}
\affiliation{%
  \institution{The Ohio State University}
  \city{Columbus}
  \state{Ohio}
  \country{USA}
}

\author{Xueru Zhang}
\affiliation{%
  \institution{The Ohio State University}
  \city{Columbus}
  \state{Ohio}
  \country{USA}}
\email{zhang.12807@osu.edu}

\renewcommand{\shortauthors}{Xie et al.}

\begin{abstract}
When machine learning (ML) algorithms are deployed to automate human-related decisions, human agents may learn the underlying decision policies and adapt their behavior strategically to secure favorable outcomes. Strategic Classification (SC) has emerged as a framework for studying this interaction between agents and decision-makers, with the goal of designing more trustworthy ML systems. 
Notably, prior theoretical models in SC typically assume that agents are perfectly or approximately rational and respond to decision policies by optimizing their utility. However, the growing prevalence of large language models (LLMs) raises the possibility that real-world agents may instead rely on these tools for strategic advice. This shift prompts two key questions: (i) Can LLMs generate effective and socially responsible strategies in SC settings to assist their users? (ii) Can existing SC theoretical models accurately capture agent behavior when agents follow LLM-generated advice in response to a decision-maker’s policy? To investigate these questions, we examine five critical SC scenarios: hiring, loan applications, school admissions, personal income, and public assistance programs. We simulate agents with diverse profiles who interact with three state-of-the-art commercial LLMs (GPT-4o, GPT-4.1, and GPT-5), following their suggestions on how to allocate effort across different features. We then compare the resulting agent behaviors with the best responses predicted by existing SC models. Our findings show that: (i) Even without access to the decision policy, LLMs can generate effective strategies that improve both scores and qualification outcomes for users; (ii) At the population level, LLM-guided effort allocation strategies yield similar or even higher score improvements, qualification rates, and fairness metrics as those predicted by the SC theoretical model, suggesting that the theoretical model may still serve as a reasonable proxy for LLM-influenced behavior; and (iii) At the individual level, LLMs tend to produce more diverse and balanced effort allocations than theoretical models, highlighting the need for developing more personalized approaches in strategic learning for real-world deployment.
\end{abstract}




\maketitle
\section{Introduction}\label{sec:intro}

Individuals subject to algorithmic decisions often adapt their behaviors strategically to the decision rule to receive desirable outcomes. As machine learning (ML) is increasingly used to make decisions about humans, there has been a growing interest in developing ML methods that explicitly consider the strategic behavior of human agents. A line of research known as \textit{strategic classification} (SC) studies this problem, in which individuals can modify their features at costs to receive favorable predictions. Recent literature on SC has focused on various aspects, including (i) designing decision policies robust to strategic behaviors \citep{Hardt2016a, Dong2018, Braverman2020, ahmadi2021strategic, horowitz2023causal}; (ii) statistical learnability of SC \citep{Sundaram2021, LevanonR22, shao2024strategic}; (iii) designing decision policies to incentivize benign strategic behaviors in SC (i.e., strategic improvement to pass an exam) \citep{Kleinberg2020, raab2021unintended, xie2024algorithmic}. Despite the diversity of previous works, most of them formulate SC as a Stackelberg game between decision-maker and human agents, where the decision-maker first publishes a policy and the agents best respond by maximizing their utilities, as shown in Def. \ref{def:sc_game}.

\begin{definition}[\citet{Hardt2016a}]\label{def:sc_game}
Consider a population of agents with feature space  $\mathcal{X}$ and label space $\mathcal{Y}=\{0,1\}$. Suppose the label is given by $y=\mathbf{1}(h(x) \geq 0.5)$ for some ground-truth labeling function
$h: \mathcal{X} \to [0,1]$. 
Let $\mathcal{D}$ denote the probability distribution over $\mathcal{X}$, and let $c: \mathcal{X} \times \mathcal{X} \to \mathbb{R}$ be a cost function that measures the cost incurred by an agent to modify their features.
\begin{enumerate}
\item The decision-maker (who knows the cost function $c$, the distribution $\mathcal{D}$, and the  labeling function $h$) publishes decision policy $\widehat{y}(x)=\mathbf{1}(f(x) \geq 0.5)$ for some scoring function $f: \mathcal{X} \to [0,1]$.  
\item  The agents (who know $c,h,\mathcal{D}$, and $f$) with features $x \in \mathcal{X}$ modify their features to $\Delta(x) \in \mathcal{X}$ at cost according to a best response function $\Delta: \mathcal{X} \to \mathcal{X}$.
\end{enumerate}
Here, $\Delta(x) = \arg\max_{z \in \mathcal{X}} f(z) - c(x,z)$ is the best response of an agent with feature $x$, where  $c(x,z)$ represents the incurred cost to modify features from $x$ to $z$. Given a loss function $\ell(\widehat{y}(x),y)$,  the decision-maker aims to find the optimal classifier $f^{*} = \arg\min_{f} \mathbb{E}_{x \sim \mathcal{D}}[\ell\bigl(\widehat{y}(\Delta(x)), y\bigr)]$. 
\end{definition}

Def. \ref{def:sc_game} specifies that agents with feature $x$ best respond to the decision policy according to $\Delta(x)$, a formulation widely adopted in prior work \citep{Dong2018, Braverman2020, levanon2021, horowitz2023causal, bechavod2022information}. Typically, the classifier $f$ belongs to a linear family (e.g., logistic regression or linear regression) and the cost function $c(x,z)$ is defined as a distance metric (e.g., the squared $l_2$ norm) measuring the distance between $x$ and $z$. Note that this form of $\Delta(x)$ assumes that agents have full knowledge of the policy $f$ and behave as perfectly rational utility maximizers. Some recent studies have relaxed these assumptions to model ``approximate" best response, e.g., by adding noise to $f$ \cite{Hardt2021}, using probability distributions to model agent behavior \cite{zhang2022}, or assuming agents first estimate a complex decision policy and then best respond to the estimated policy \cite{ghalme2021strategic, xie2024nonlinearwelfareawarestrategiclearning}.

 
While theoretical frameworks offer an elegant and rigorous approach to studying the SC problem, the rapid development of large language models (LLMs) introduces new challenges for modeling best responses in strategic learning settings. Several recent studies have shown that LLMs can assist humans in making complex decisions, including (i) economic and marketing decisions \cite{immorlica2024generative}; (ii) completing surveys with realistic responses \cite{dominguez2023questioning}; (iii) modeling patients with mental health issues \cite{wang2024patient}; and (iv) simulating human trust behaviors \cite{xie2024can}. Given that commercial LLMs have already attracted hundreds of millions of users worldwide \cite{backlinko_chatgpt_stats}, it is increasingly common and reasonable for human agents to consult LLMs before responding to decision policies in real-world SC scenarios. For example, applicants may seek advice from tools such as  GPT-4o \cite{achiam2023gpt} or Claude 3 \cite{claude} to learn concrete strategies for improving their profiles to obtain loan approvals, college admissions, or job offers. These developments raise two natural questions: (i) Can LLMs produce effective and socially responsible strategies in SC settings that benefit their users? (ii) Can the theoretical model described above still approximate agent behavior when agents respond according to LLM-generated advice? 

To answer these questions, we conduct simulation studies in which strategic behaviors are generated either by best response functions from prior literature or by strategies produced by $3$ state-of-the-art commercial LLMs: GPT-4o \cite{gpt4o}, GPT-4.1\cite{openai2025gpt41}, and GPT-5 \cite{gpt5}. We then compare the results across these models. Specifically, we examine five well-known settings where strategic classification is known to occur: (i) loan applications \cite{perdomo2020}; (ii) personal income prediction \cite{ding2021retiring}; (iii) law school admissions \cite{lawdata}; (iv) public assistance programs \cite{ding2021retiring}; and (v) hiring \cite{hiringdata}. In each setting, we assume that the decision-maker first publishes a decision policy (which may differ from the true labeling function), based on which strategic human agents allocate effort to modify their features. We further assume that agents do not have direct access to the policy and instead consult LLMs for guidance on how to allocate their effort. After agents respond to the policy, we compare the resulting changes in their scores (i.e., decision outcomes), true qualifications (i.e., labels), and the unfairness induced by their effort allocation strategies with those produced by the theoretical best response model defined in Def. \ref{def:sc_game}. Our findings can be summarized as follows:

\begin{enumerate}
    \item All LLMs can generate effort allocation strategies that match the performance of the theoretical model in most settings. At the population level, all LLMs produce effort allocation strategies that, \textbf{on average}, lead to similar or even higher score increases and qualification improvements as those predicted by the theoretical model. GPT-5 even outperforms the theoretical model in $2$ settings.

    \item At the individual level, LLMs tend to generate more diverse effort allocation strategies for users with distinct features. Additionally, the strategies produced by LLMs are more balanced, with efforts distributed more evenly across features.

    \item The average effort allocations generated by GPT-5 exhibit stronger alignment with the theoretical model. Also, the average effort allocations generated by LLMs more closely align with the theoretical model in decision settings with fewer features.

    \item The effort allocation strategies generated by all LLMs achieve similar levels of fairness as the theoretical model, and the overall unfairness is not significant.
\end{enumerate}

The first and second findings together suggest that the theoretical model can serve as a reasonable proxy for LLM-generated responses, as both result in comparable score increases, qualification improvements, and fairness. However, the third finding highlights the more complex nature of LLM-generated responses compared to the theoretical model, indicating that a more individualized approach may be needed for modeling agent responses in SC settings.

\subsection{Related Work}\label{subsec:related}
Our paper is closely related to two lines of prior work, which we discuss below.

\subsubsection{Strategic classification.} Strategic classification (SC) was first modeled by \citet{Hardt2016a} to demonstrate the interaction between individuals and a decision maker as a Stackelberg game. By incorporating individuals' best responses, the decision maker can optimize decisions by anticipating strategic behavior. In recent years, more sophisticated models of strategic classification have emerged \citep{Ben2017, Dong2018, Braverman2020, Hardt2021, Izzo2021, ahmadi2021strategic, tang2021, zhang2020, zhang2022, Eilat2022, liu22, lechner2022learning, chen2020learning, xie2024nonlinearwelfareawarestrategiclearning, xie2024learning}. For example, \citet{Ben2017} developed a linear regression predictor in a competitive setting, where two players aim to outperform each other in prediction accuracy. \citet{Dong2018} extended this to the online version of the strategic classification algorithm. \citet{chen2020learning} introduced a strategic-aware linear classifier aimed at minimizing Stackelberg regret. \citet{tang2021} considered scenarios where the decision maker is only aware of a subset of individuals' actions. \citet{LevanonR22} expanded the framework to cases where individuals and the decision maker share aligned interests. \citet{lechner2022learning} proposed a new loss function balancing prediction accuracy with resistance to strategic manipulation, while \citet{Eilat2022} relaxed the assumption of independent individual responses, presenting a robust learning framework utilizing a Graph Neural Network. \citet{xie2024nonlinearwelfareawarestrategiclearning} developed a framework for welfare-aware strategic learning, and \citet{gemalmaz2024understanding} explored how fairness influences the strategies of human agents. Meanwhile, SC is also closely related to algorithmic recourse, with \citet{karimi2022survey} providing a comprehensive survey on the topic. Note that it remains an open question how the theoretical best response model in SC fits in reality. Research suggests that real-world strategic responses are often more nuanced and complex than these frameworks account for \cite{ebrahimi2024double}. Existing studies in SC are validated using static datasets or simulating dynamic data resulting from agent responses using pre-defined rules \cite{simulate2020,miller2020whynot}. In our paper, we mainly focus on comparing the LLM-generated responses to the theoretical model, while it remains an interesting direction to conduct human subject studies to compare LLM-generated responses and real human responses.

\subsubsection{LLMs can model real-world human behaviors.} A growing body of recent literature has explored the use of LLMs to complete complex tasks in social science that require reasoning similar to human behavior \cite{engel2024integrating}. In addition to the examples mentioned above, \citet{brand2023using} examined whether LLMs can conduct market research, while \citet{shapira2024can} focused on whether LLMs can simulate human choices in economic markets. \citet{taitler2024braess} further investigated how LLMs can serve as a platform for revenue maximization. \citet{hartmann2023political} demonstrated the political opinions of ChatGPT, and \citet{ziems2024can} discussed the transformative potential of LLMs for computational social science. \citet{li2023camel} highlighted that LLMs can play roles and solve complex tasks in alignment with these roles, while \citet{argyle2023out} studied how LLMs can simulate human agents within different social groups. Collectively, these works suggest that LLMs have significant potential to simulate real-world human agents. \citet{park2024llm} examined whether LLMs demonstrate regret in online learning games. For more on how LLMs can act as strategic human agents, we refer readers to a recent survey paper \citep{zhang2024llm}.

\section{Simulation Settings}\label{sec:setting}

\subsection{Problem Formulation}\label{subsec:problem}

In each of the five settings, we construct a binary classification task with feature space $\mathcal{X} \subseteq \mathbb{R}^d$ and label space $\mathcal{Y} = \{0, 1\}$. Assuming all features are drawn from a distribution $\mathcal{D}$, we first train a labeling function $h: \mathcal{X} \rightarrow [0,1]$ on a training dataset and define the ground-truth label based on the threshold classifier $\mathbf{1}(h(x) \ge 0.5)$. 

\paragraph{Decision-maker's policy.} We assume that the decision-maker trains a policy $\mathbf{1}(f(x) \ge 0.5)$ using the same training dataset, where $f: \mathcal{X} \rightarrow [0,1]$ is a scoring function. Note that $f$ and $h$ are not necessarily the same because they can come from different function families. For example, $h$ may be a highly complex neural network, while $f$ could be a simple logistic function. We consider two scenarios: (i) labeling function $h$ is relatively simple, in which case the decision-maker publishes $f = h$; (ii) when $h$ is complex, the decision-maker may instead publish a simpler scoring function $f$, due to constraints such as limited training data, legal requirements for transparency, or the need for model explainability \cite{Rose2020, xie2024nonlinearwelfareawarestrategiclearning}.

\paragraph{Theoretical model for agent responses.} Given the decision policy $f$, we sample $1000$ human agents with features $x \sim \mathcal{D}$ to participate in the decision system, where each agent strategically responds to the published policy $f$. We then model agent responses using a generalized form of the setup defined in Def. \ref{def:sc_game}.

Specifically, we assume that agents can strategically modify a subset of their features by investing efforts $ e\in {\mathbb {R}_+^d}$ subject to a quadratic cost $\frac{1}{2}||e||^2$. We further assume that modifying each of the $d$ features incurs a different effort-to-feature conversion ratio, represented by an effort conversion matrix $W = \text{diag}(w) \in \mathbb{R}^{d \times d}$, where $w \in \mathbb{R}^d$.\footnote{The vector $w$ captures the varying difficulty of modifying different features. For example, in a loan application scenario, it may be easier for an applicant to apply for more credit cards than to increase their monthly income. The vector $w$ is set according to the practical difficulty of modifying each feature in each setting.} Formally, let $x(e)$ denote the modified feature vector after effort $e$, then we have $x(e) = x + We$. Additionally, we assume that agents with features $x$ use a \textit{first-order approximation} to estimate the decision policy $f(x')$, defined as $$Q(x') = x + \frac{\nabla f(x)^T}{\|\nabla f(x)\|} (x' - x).$$ This approximation is motivated by the fact that human agents may struggle to effectively learn $f$ when policies are highly complex and non-linear \cite{xie2024nonlinearwelfareawarestrategiclearning}. We normalize $\nabla f(x)$ to a unit vector to maintain consistency with prior literature \cite{bechavod2022information} and to prevent overly large gradient values.

Compared to the original setup in Def.~\ref{def:sc_game}, the above formulation better reflects human behavior in practice and is a more general framework. Specifically, when interacting with LLMs, human agents are more likely to seek guidance on how to allocate effort across features, rather than directly requesting modified feature values. Moreover, the effort vector $e$ and the conversion matrix $W$ together play a role analogous to the cost function $c$ in Def.~\ref{def:sc_game}. We provide a formal proof in 
Appendix~\ref{app:proof} demonstrating the equivalence between this formulation and Def. \ref{def:sc_game} under the assumption of a linear decision policy and a quadratic cost function---conditions commonly assumed in prior SC literature (e.g., \citet{Braverman2020, zhang2022, xie2024automating}). Formally,  agent responses in our theoretical model are defined as follows.  


\begin{definition}[Theoretical agent response]\label{def:br}
    Agents with initial features $x$ will strategically modify their features to $x^{*}$ as follows.
    \begin{align*}
        x^{*} = \max_{e \in {\mathbb R^d_{+}}} Q\left(x(e)\right) - \frac{1}{2}||e||^2
    \end{align*}
\end{definition}

\paragraph{Two types of features.} Similar to prior work on causal strategic learning \cite{Miller2020,shavit2020,horowitz2023causal}, we assume that features can be \textit{causal }(i.e., feature changes result in the label changes) or \textit{non-causal} (i.e., feature changes do not affect the label). This concept is closely related to social welfare in SC \cite{bechavod2022information, xie2024nonlinearwelfareawarestrategiclearning}, where modifying causal features reflects genuine qualification improvement, while modifying non-causal features corresponds to strategic manipulation or gaming. In our framework, this means that changes to causal features alter the ground-truth labeling function $h(x)$, whereas changes to non-causal features do not affect $h(x)$, though they may still influence the scoring function $f(x)$. Additionally, we assume that human agents need to invest more effort per unit change in causal features (i.e., the corresponding values in $w$ are larger), as genuine improvement is typically more difficult than manipulation \cite{Barsotti2022, Jin2022}.

\begin{figure}
    \centering
    \includegraphics[width=0.3\textwidth]{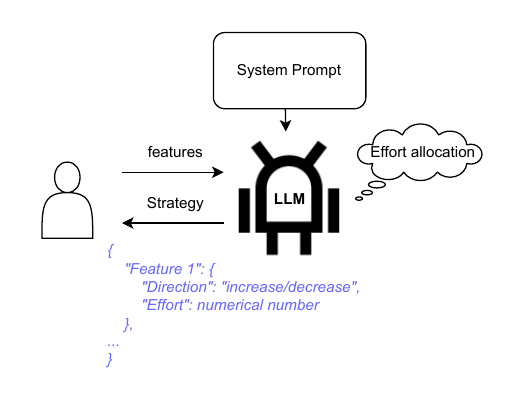}
    \includegraphics[width=0.69\textwidth]{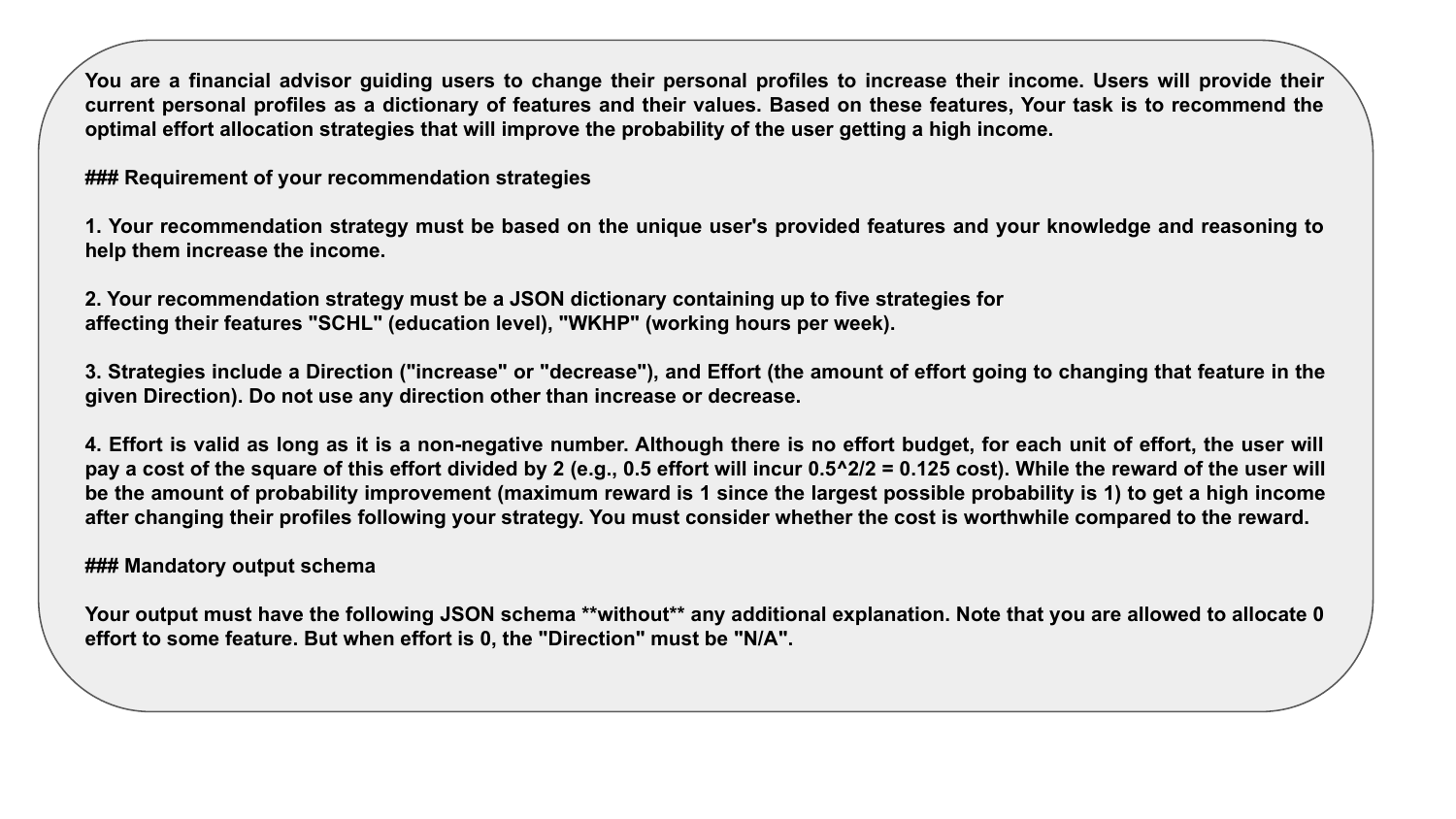}
    \caption{An illustration of how human agents seek advice from LLMs in strategic classification (SC) settings (left), and an example prompt used in the personal income setting (right).}
    \label{fig:prompt}
\end{figure}

\subsection{Datasets}\label{subsec:data}

We consider five different yet important strategic classification (SC) settings and describe each in detail below.

\paragraph{Hiring \cite{hiringdata}.} 
This dataset contains over 70,000 records of job applicants and their hiring outcomes for technical positions. The analysis considers four strategically modifiable features, two of which are causal: education level (\texttt{Education}) and the number of computer skills (\texttt{ComputerSkills}). The remaining two---previous job salary (\texttt{PreviousSalary}) and self-reported years of coding experience (\texttt{YearCode})---are non-causal. Additionally, age (\texttt{age}) is included as a sensitive, unmodifiable feature. The model assumes the effort conversion vector is $w = [1,1,2,2]$.

\paragraph{Personal Income \cite{ding2021retiring}.} 
The dataset includes records of over 190,000 individuals, with the target variable indicating whether a citizen's income exceeds \$50,000. The model considers two strategically modifiable features, both of which are causal: education level (\texttt{SCHL}) and weekly work hours (\texttt{WKHP}). Age (\texttt{age}) is included as a sensitive, unmodifiable feature. The effort conversion vector is set to $w = [0.5, 1]$.

\paragraph{Law School \cite{lawdata}.} 
This dataset encompasses 27,000 law student records from 1991 to 1997, with bar exam passage as the target variable. The analysis considers two strategically modifiable features, both of which are causal: undergraduate GPA (\texttt{UGPA}) and LSAT score (\texttt{LSAT}). Sex (\texttt{sex}) is incorporated as a sensitive, unmodifiable feature. The effort conversion vector is $w = [0.5,0.5]$.

\paragraph{Loan Approval \cite{creditdata}.} 
It contains 250,000 borrower records, with loan repayment as the target variable. The analysis considers five strategically modifiable features. Two of which are causal: debt ratio (\texttt{\seqsplit{Debt Ratio}}) and monthly income (\texttt{\seqsplit{Monthly Income}}). The remaining three features are non-causal: revolving utilization ratio (\texttt{\seqsplit{RevolvingUtilizationOfUnsecuredLines}}), number of open credit lines (\texttt{\seqsplit{NumberOfOpenCreditLinesAndLoans}}), and number of real estate loans (\texttt{\seqsplit{NumberRealEstateLoansOrLines}}). Age (\texttt{age}) is included as a sensitive, unmodifiable feature. The effort conversion vector is $w = [0.5,0.5,2,2,1]$.

\paragraph{Public Assistance Program \cite{ding2021retiring}.} 
This dataset contains records of over 190,000 citizens, with program qualification as the target variable. The analysis considers three strategically modifiable features, two of which are causal: education level (\texttt{SCHL}) and total income (\texttt{PINCP}). The third feature---weekly work hours (\texttt{WKHP})---is non-causal. Age (\texttt{age}) serves as a sensitive, unmodifiable feature. The effort conversion vector is $w = [1,1,2]$.

\subsection{Seek advice from LLMs.} 

We consider a realistic scenario in which human agents lack quantitative knowledge of the decision function $f$. Instead, they seek advice from knowledgeable ``experts" (i.e., LLMs) and follow their suggestions to modify their features. Specifically, for each of the five settings, we first construct prompts that describe the task and then instruct the LLMs to act as advisors, helping users achieve their desired decision outcomes while accounting for the trade-off between costs and rewards.  The right panel of Fig. \ref{fig:prompt} illustrates the prompt used for the personal income setting; prompts for the remaining settings are constructed similarly and provided in Appendix \ref{app:prompt}. For ease of comparison, we constrain the LLMs to output only the effort allocation for each user, without textual explanation. These outputs include only the magnitude and direction of efforts allocated to each feature. To ensure the outputs are in the same format, we also specify the JSON format as shown in the left panel of Fig. \ref{fig:prompt}.

We generate results using gpt-4o-2024-10-26~\cite{gpt4o}, gpt-4.1-2025-04-14~\cite{openai2025gpt41}, and gpt-5-2025-08-07~\cite{gpt5}. When the option is available, we set the temperature to $0$ to elicit the most optimal answers from the LLMs' perspective. All other hyperparameters are set to their default values.

\section{Main Results: Do LLMs Align with the Theoretical Model?}

For each of the five settings above, we assume $1000$ human agents participate in the decision system and strategically respond to the deployed policy $f$. We first obtain the theoretical best responses and the agent responses produced by LLMs, and then provide a detailed comparison analysis. Specifically, we compare strategies produced by LLMs and the theoretical models in the following four aspects: (i) the distributions of effort allocation strategies in each setting; (ii) the resulting score increase (Def. \ref{def:si}); (iii) the resulting qualification improvement (Def. \ref{def:qi}); and (iv) the unfairness (Def. \ref{def:uf}) resulting from the efforts. We first provide the relevant definitions as follows.

\begin{definition}[Score increase]\label{def:si} Let $x$ and $x'$ be the features of an agent before and after responding to the deployed policy $f$, respectively. The resulting score increase is measured as $f(x')-f(x)$.
\end{definition}

\begin{definition}[Qualification improvement]\label{def:qi} Let $x_c$ be the subset of causal features that causally affect the agents' labels. For an agent whose features change from $x$ to $x'$, its qualification improvement is measured as $h(x'_c)-h(x_c)$.
\end{definition}

\begin{definition}[Unfairness of effort allocation]\label{def:uf}
    For each of the five settings, there is a sensitive feature that divides the human agents into distinct social groups. The unfairness resulting from an effort allocation is measured by the difference in the average score increase or qualification improvement between these social groups.
\end{definition}

The score increase measures the effectiveness of the agent responses to the decision-maker, while the qualification improvement reflects the social welfare resulting from the deployment of policy $f$. In our experiments, we assume that the deployed policy $f$ is always a logistic classifier. The ground-truth labeling function $h$ can either be simple (identical to $f$, \textbf{decision scenario 1}) or complex (an MLP classifier, \textbf{decision scenario 2}). For each setting, we present experimental results for all GPT-4o, GPT-4.1, and GPT-5.

\begin{figure}[h]
    \centering
    \includegraphics[width= 0.6\textwidth]{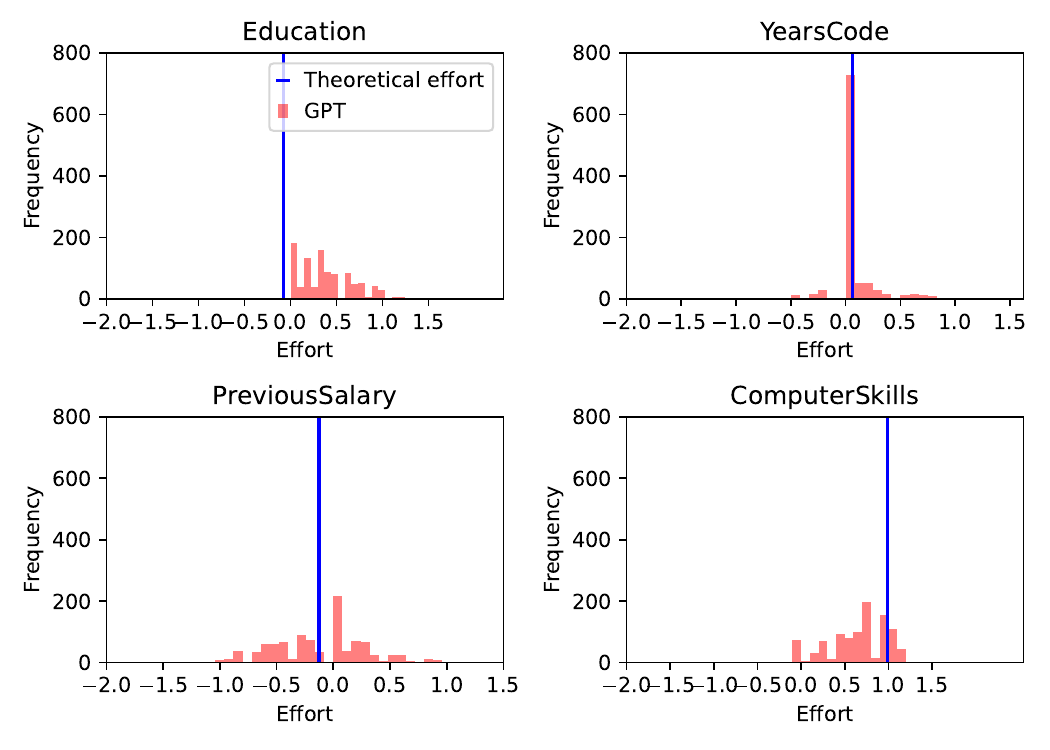}
    \caption{Comparison of effort allocations between the theoretical model and LLM in the \textbf{hiring} setting, produced by GPT-5. As the decision policies are identical in both decision scenarios, the resulting effort allocations are also the same.}
    \label{fig:effort_hiring_d12_5}
\end{figure}

\begin{figure}[h]
    \centering
    \includegraphics[width= 0.6\textwidth]{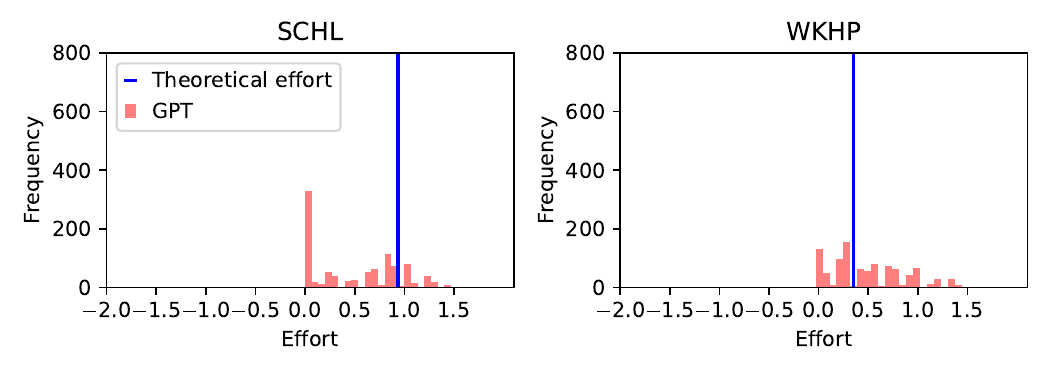}
    \caption{Comparison of effort allocations between the theoretical model and LLM in the \textbf{income} setting, produced by GPT-5. As the decision policies are identical in both decision scenarios, the resulting effort allocations are also the same.}
    \label{fig:effort_income_d12_5}
\end{figure}

\begin{figure}[h]
    \centering
    \includegraphics[width= 0.6\textwidth]{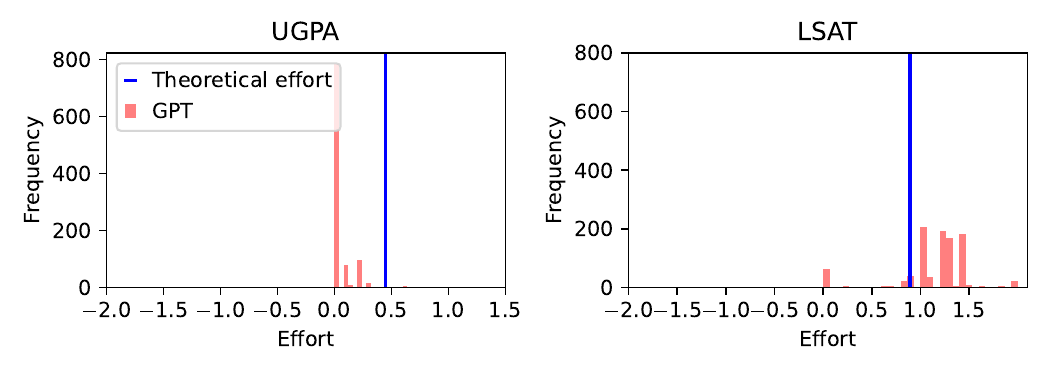}
    \caption{Comparison of effort allocations between the theoretical model and LLM in the \textbf{law school} setting, produced by GPT-5. As the decision policies are identical in both decision scenarios, the resulting effort allocations are also the same.}
    \label{fig:effort_law_d12_5}
\end{figure}

\begin{figure}[h]
    \centering
    \includegraphics[width= 0.9\textwidth]{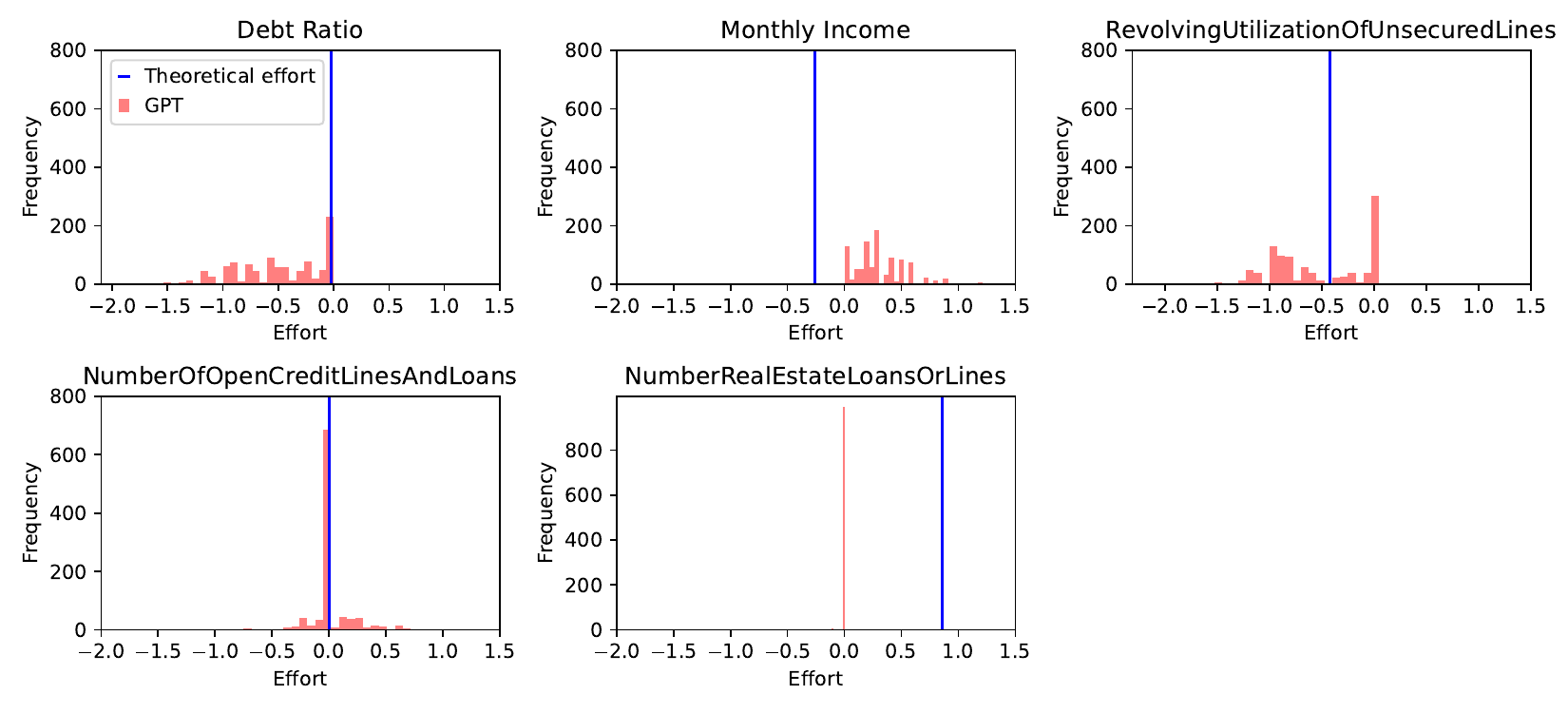}
    \caption{Comparison of effort allocations between the theoretical model and LLM in the \textbf{loan approval} setting, produced by GPT-5. As the decision policies are identical in both decision scenarios, the resulting effort allocations are also the same.}
    \label{fig:effort_credit_d12_5}
\end{figure}

\begin{figure}[h]
    \centering
    \includegraphics[width= 0.85\textwidth]{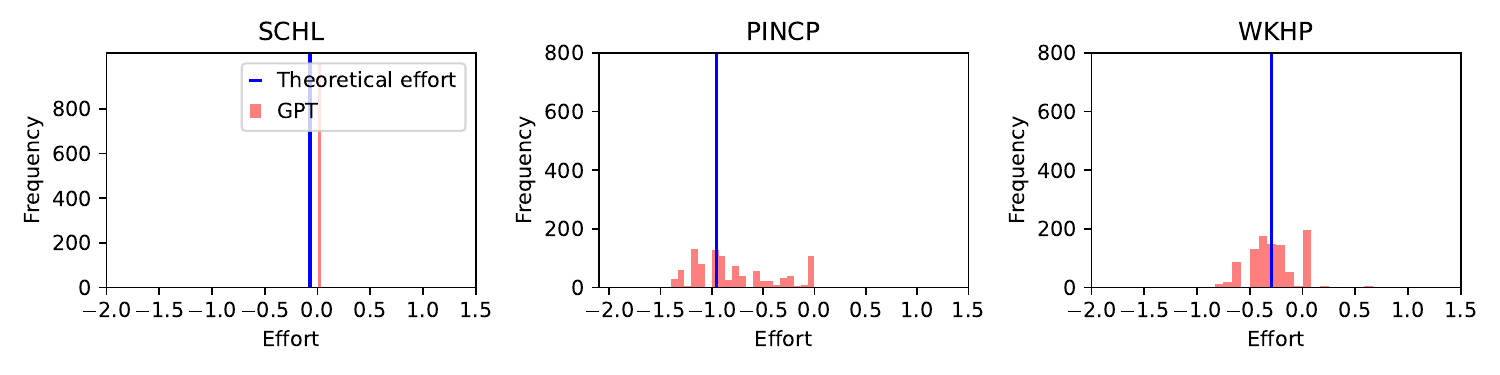}
    \caption{Comparison of effort allocations between the theoretical model and LLM in the \textbf{public assistance} setting, produced by GPT-5. As the decision policies are identical in both decision scenarios, the resulting effort allocations are also the same.}
    \label{fig:effort_pap_d12_5}
\end{figure}

\subsection{Effort Allocation Comparisons}\label{subsec:ea}

Next, we illustrate the effort allocation strategies for LLMs and the theoretical model, as shown in Fig. \ref{fig:effort_hiring_d12_5} through Fig. \ref{fig:effort_pap_d12_5} where the results of GPT-5 are presented. For the plots of GPT-4o and GPT-4.1, we defer them to Fig. \ref{fig:effort_hiring_d12} to \ref{fig:effort_pap_d12} in App. \ref{app:add_exp}. The red bars represent the effort allocation distributions generated by the LLMs, while the blue lines are those produced by the theoretical model. Since the decision-maker employs a logistic classifier, which belongs to the linear model family, the theoretical best responses (Def. \ref{def:br}) remain the same across all feature values\footnote{Although the uniform theoretical responses may appear simplistic, remain the most widely used model in strategic learning literature (e.g., \citet{Hardt2016a, bechavod2022information, xie2024automating}). Some works introduce noise into the response (e.g., \citet{Hardt2021}) but the agent responses are still centered around a single point.}. 

\textbf{Overall, LLMs tend to produce more diverse and balanced effort allocations compared to the theoretical model.} The effort distributions generated by LLMs (red bars) are not always centered at the effort value of the theoretical model (blue line), indicating that LLMs produce distinct effort allocation strategies for users with different features. In $4$ of the $5$ settings, the average effort allocation of GPT-5 has lower variance than the theoretical model's and LLM-produced effort allocations always achieve the lowest variance in all $5$ settings, as shown in Tab. \ref{tab:feature_allocation}. For example, in the hiring setting, the theoretical model tends to allocate most of the effort on \texttt{ComputerSkills}, while LLMs distributes efforts more evenly between \texttt{ComputerSkills} and \texttt{Education}. In the public assistance program setting, the theoretical model almost focuses solely on \texttt{PINCP} (i.e., manipulating personal income to be lower), whereas  LLMs distribute efforts across all three features.  

\textbf{Meanwhile, the consistency between the average of LLM-induced effort allocations and the theoretical model is higher when the decision environment involves fewer features.} We report the average effort allocated to each feature by both LLMs and the theoretical model in Tab. \ref{tab:feature_allocation} and show the $l_2$ distances between the effort allocation vectors produced by LLMs and the theoretical model in the last rows in each cell. They show that LLMs produce effort allocations closely aligned with the theoretical model in the law school setting with only two features, whereas their allocations in the loan approval setting deviate substantially from the theoretical model when five features are involved. More importantly, \textbf{the theoretical model seems to align more with GPT-5.}  We observe that GPT-5 produces effort allocations closest to the theoretical model in $3$ out of the $5$ settings, suggesting that the most recent LLM seems to reason similarly to the theoretical model.

\begin{table*}[h!]
\centering
\caption{Average efforts (averaged on all agents) allocated to each feature in each setting produced by GPT-4o, GPT-4.1, and the theoretical model. Specifically, the average effort allocations produced by the theoretical model (the last column) are more polarized than the ones produced by LLMs.}
\label{tab:feature_allocation}
\resizebox{0.8\textwidth}{!}{
\begin{tabular}{llcccc}
\toprule
\textbf{Setting}           & \textbf{Feature}                   & \textbf{GPT-4o} & \textbf{GPT-4.1}  & \textbf{GPT-5} & \textbf{Theoretical} \\
\midrule
\multirow{4}{*}{\textbf{Hiring}}    
                            & \texttt{Education}                & $0.496$          & $0.343$   &$0.383$      & $-0.072$            \\
                            & \texttt{YearCode}                 & $0.205$          & $0.117$   &$0.047$      & $-0.126$             \\
                            & \texttt{PreviousSalary}           & $0.211$          & $0.041$    &$-0.120$      & $-0.063$            \\
                            & \texttt{ComputerSkills}           & $0.443$          & $0.309$    &$0.676$     & $0.987$             \\
                            & \textit{Variance of average effort magnitude}                 & $0.017$          & $\boldsymbol{0.016}$  & $0.094$        & $0.152$              \\
                            & \textit{$l_2$ distance to theoretical responses}
                            &$0.896$ &$0.838$ &$\mathbf{0.531}$ & / \\
                            
\midrule
\multirow{2}{*}{\textbf{Income}}     
                            & \texttt{SCHL}                     & $0.513$          & $0.513$  & $0.499$       & $0.935$             \\
                            & \texttt{WKHP}                     & $0.782$          & $0.782$    & $0.506$     & $0.356$             \\
                            & \textit{Variance of average effort magnitude}  & $0.018$     & $\boldsymbol{0.000}$     & $0.018$         & $0.084$              \\
                            & \textit{$l_2$ distance to theoretical responses}
                            &$0.599$ &$0.599$ &$\mathbf{0.461}$ & / \\
\midrule
\multirow{2}{*}{\textbf{Law School}} 
                            & \texttt{UGPA}                     & $0.345$          & $0.555$   &$0.043$      & $0.450$             \\
                            & \texttt{LSAT}                     & $0.899$          & $0.702$  &$1.115$       & $0.893$             \\
                            & \textit{Variance of average effort magnitude}                  & $0.076$          & $\boldsymbol{0.005}$   &$0.287$      & $0.049$              \\
                            & \textit{$l_2$ distance to theoretical responses}
                            &$\mathbf{0.105}$ &$0.218$ &$0.404$ & / \\
\midrule
\multirow{5}{*}{\textbf{Loan Approval}} 
                            & \texttt{Debt Ratio}               & $-0.319$         & $-0.243$   & $-0.500$      & $-0.026$            \\
                            & \texttt{Monthly Income}           & $0.351$          & $0.182$   & $0.312$       & $-0.260$            \\
                            & \texttt{RevolvingUtilizationOfUnsecuredLines} & $-0.235$ & $-0.383$  & $-0.550$       & $-0.423$            \\
                            & \texttt{NumberOfOpenCreditLinesAndLoans}    & $0.132$ & $-0.022$   & $0.025$     & $0.002$             \\
                            & \texttt{NumberRealEstateLoansOrLines}      & $0.180$ & $0.075$   & $-0.002$      & $0.861$             \\
                            & \textit{Variance of average effort magnitude}                  & $\boldsymbol{0.007}$          & $0.016$    & $0.100$      & $0.100$              \\
                            & \textit{$l_2$ distance to theoretical responses}
                            &$0.987$ &$\mathbf{0.928}$ &$1.146$ & / \\

\midrule
\multirow{3}{*}{\textbf{Public Assistance}} 
                            & \texttt{SCHL}                     & $0.227$         & $0.229$ & $0.000$       & $-0.072$            \\
                            & \texttt{PINCP}                     & $-1.043$          & $-1.041$   & $-0.804$       & $-0.953$            \\
                            & \texttt{WKHP}                    & $-0.489$         & $-0.476$   & $-0.281$     & $-0.295$            \\
                            & \textit{Variance of average effort magnitude}                  & $0.116$          & $\boldsymbol{0.115}$    & $0.111$     & $0.140$              \\
                            & \textit{$l_2$ distance to theoretical responses}
                            &$0.368$ &$0.362$ &$\mathbf{0.166}$ & / \\
\bottomrule
\end{tabular}}
\end{table*}

\subsection{Score Increase}\label{subsec:score}

In this section, we present the score increases achieved when human agents adjust their efforts based on either the theoretical model or the strategies suggested by LLMs, assuming a logistic classifier  $f$ is used as the decision policy. The average score increases for all agents are reported in the first row of each setting in Tab. \ref{tab:combined_sq_summary}. Notably, LLMs are able to match the score improvements of the theoretical model in most settings. \textbf{GPT-5 even slightly outperforms the theoretical model in $2$ settings, while GPT-4o and GPT-4.1 generally guide users to achieve average score increases only slightly below those achieved by the theoretical model, demonstrating their effectiveness in guiding effort allocation.} Note that the higher score improvements occur because, although we explicitly instruct the LLMs to consider the trade-off between reward and cost, they do not have access to the exact scoring function $f$, and may sometimes produce effort allocations where the cost exceeds the benefit.

\begin{figure*}[h]
    \centering
    \begin{subfigure}{0.32\textwidth}
        \centering
        \includegraphics[width= \linewidth]{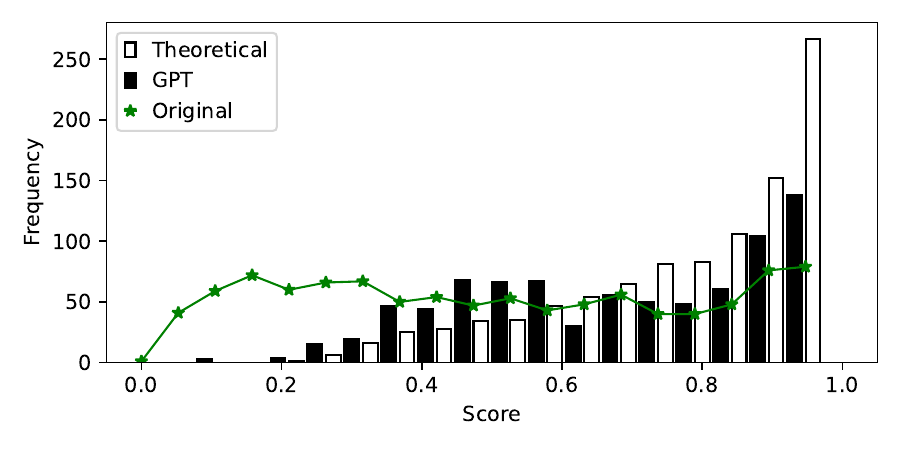} 
        \caption{GPT-4o}
        \label{subfig:score_hiring_d12_4o}
    \end{subfigure}
    \hspace{0.1cm}
    \begin{subfigure}{0.32\textwidth}
        \centering
        \includegraphics[width= \linewidth]{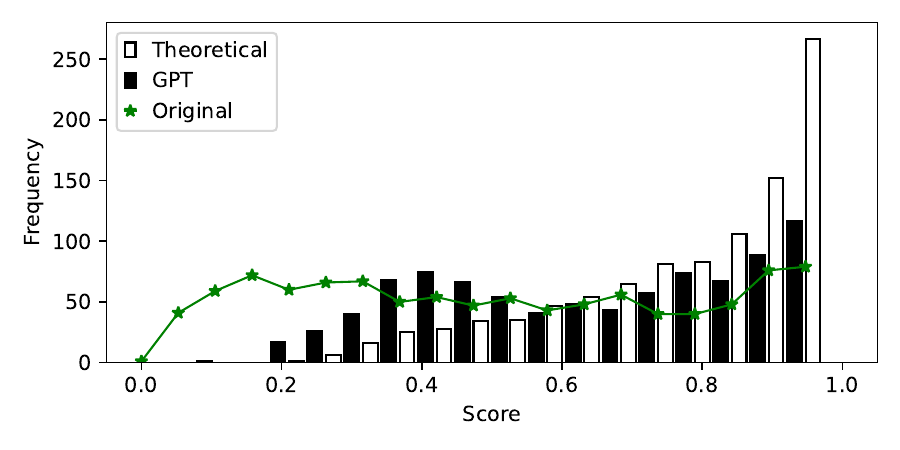}
        \caption{GPT-4.1}
        \label{subfig:score_hiring_d12_41}
    \end{subfigure}
    \hspace{0.1cm}
    \begin{subfigure}{0.32\textwidth}
        \centering
        \includegraphics[width= \linewidth]{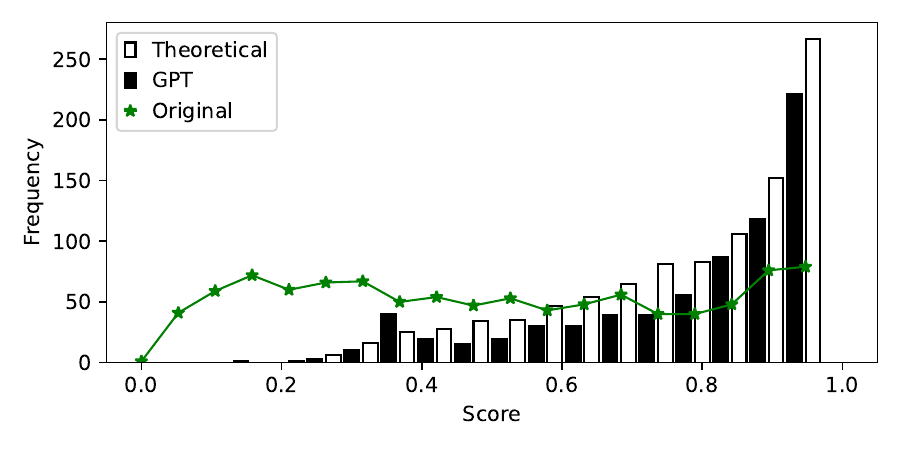}
        \caption{GPT-5}
        \label{subfig:score_hiring_d12_5}
    \end{subfigure}
    \caption{Comparison of the score distribution before and after the agent's response in the \textbf{hiring} setting: while GPT-4o and GPT-4.1 achieve smaller score increases than the theoretical model (as indicated by the rightward skew of the hollow bars in the first $2$ plots), GPT-5 achieves slightly higher score increase (solid bars in the rightmost plot) than the theoretical model.}
    \label{fig:score_hiring_d12}
\end{figure*}

Additionally, we visualize the score distributions of agents before responding, after responding based on LLM guidance, and after responding based on the theoretical model across all five settings. Fig.~\ref{fig:score_hiring_d12} presents the visualization for the hiring setting, with the remaining plots provided in the Appendix (Fig.~\ref{fig:score_income_d1} through \ref{fig:score_pap_d1}). In each plot, hollow bars represent post-response score distributions under the theoretical model, while solid bars represent those guided by LLMs. The results show that LLMs effectively provide strategies to help agents improve their scores, i.e., solid bars are right-skewed compared to the green lines.

\begin{figure*}[h]
    \centering
    \begin{subfigure}{0.32\textwidth}
        \centering
        \includegraphics[width= \linewidth]{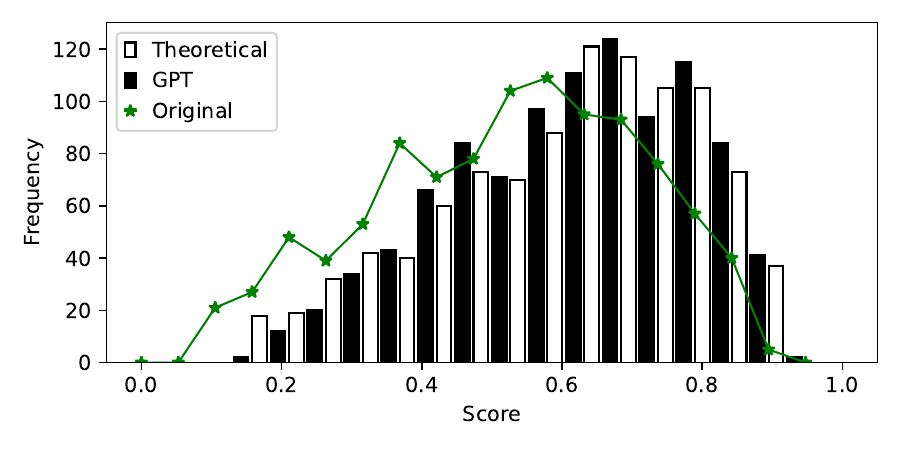}
        \caption{GPT-4o}
        \label{subfig:q_law_d1_4o}
    \end{subfigure}
    \hspace{0.1cm}
    \begin{subfigure}{0.32\textwidth}
        \centering
        \includegraphics[width= \linewidth]{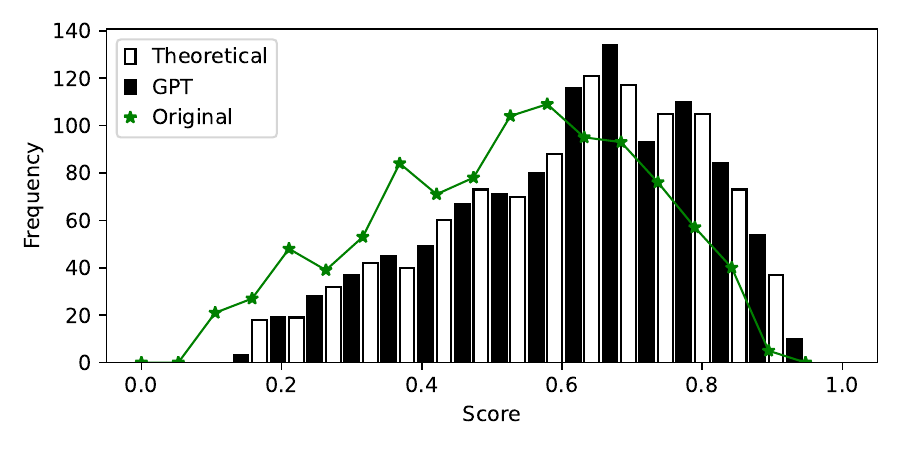}
        \caption{GPT-4.1}
        \label{subfig:q_law_d1_41}
    \end{subfigure}
    \hspace{0.1cm}
    \begin{subfigure}{0.32\textwidth}
        \centering
        \includegraphics[width= \linewidth]{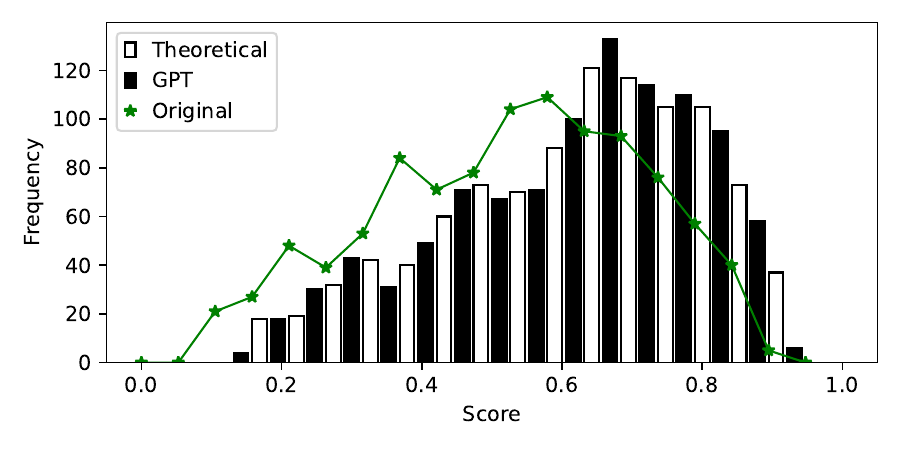}
        \caption{GPT-5}
        \label{subfig:q_law_d1_5}
    \end{subfigure}
    \caption{Comparison of the qualification distribution before and after agent responses in the \textbf{law school} setting for decision scenario 1. LLMs (the solid bars) achieve comparable qualification improvements as the theoretical model (the hollow bars).}
    \label{fig:q_law_d1}
\end{figure*}
\begin{table*}[h!]
\centering
\caption{Score increase and qualification improvement for LLMs and the theoretical model across different settings under decision scenario 1 (S1) and decision scenario 2 (S2).}
\label{tab:combined_sq_summary}
\resizebox{0.75\textwidth}{!}{
\begin{tabular}{llcccc}
\toprule
\textbf{Setting} & \textbf{Metric}                           & \textbf{Theoretical} & \textbf{GPT-4o} & \textbf{GPT-4.1} & \textbf{GPT-5} \\
\midrule
\multirow{3}{*}{\textbf{Hiring}} 
    & Score increase                     & $0.273$            & $0.212$          & $0.160$     & $\mathbf{0.312}$     \\
    & Qualification improvement (S1)     & $0.269$            & $0.269$         & $0.160$      & $\mathbf{0.308}$   \\
    & Qualification improvement (S2)     & $0.279$            & $0.230$         & $0.177$      & $\mathbf{0.303}$   \\
\midrule
\multirow{3}{*}{\textbf{Income}} 
    & Score increase                     & $\mathbf{0.023}$            & $0.017$          & $0.017$     &$0.014$     \\
    & Qualification improvement (S1)     & $\mathbf{0.023}$            & $0.017$         & $0.017$      &$0.014$   \\
    & Qualification improvement (S2)     & $\mathbf{0.023}$            & $0.022$         & $0.022$      &$0.015$   \\
\midrule
\multirow{3}{*}{\textbf{Law school}} 
    & Score increase                     & $0.081$            & $0.072$          & $0.078$     & $\mathbf{0.083}$      \\
    & Qualification improvement (S1)     & $0.081$            & $0.072$          & $0.078$     & $\mathbf{0.083}$     \\
    & Qualification improvement (S2)     & $0.070$            & $0.063$          & $0.066$     & $\mathbf{0.071}$    \\
\midrule
\multirow{3}{*}{\textbf{Loan approval}} 
    & Score increase                     & $\mathbf{0.095}$& $0.010$          & $0.007$     & $0.002$    \\
    & Qualification improvement (S1)     & $\mathbf{0.001}$            & $-0.011$         & $-0.008$   & $-0.002$      \\
    & Qualification improvement (S2)     & $0.025$            & $0.003$         & $0.038$    &$\mathbf{0.092}$     \\
\midrule
\multirow{3}{*}{\textbf{Public assistance program}} 
    & Score increase                     & $0.215$            & $\mathbf{0.219}$         & $0.220$ & $0.179$         \\
    & Qualification improvement (S1)     & $0.198$            & $0.208$         & $\mathbf{0.210}$  & $0.170$        \\
    & Qualification improvement (S2)     & $\mathbf{0.201}$            & $0.171$         & $0.173$    & $0.173$       \\
\bottomrule
\end{tabular}
}
\end{table*}

\subsection{Agent Qualification Improvement}\label{subsec:qualification}

Next, we compare the qualification improvements achieved by agents under effort allocation strategies derived from the theoretical model versus those generated by LLMs. Specifically, we evaluate these strategies under two distinct decision scenarios, defined as follows.

\paragraph{Decision scenario 1.} 
This scenario occurs when the decision-maker directly deploys $f=h$, corresponding to the classic setup in strategic classification (e.g., \citet{Hardt2016a, Dong2018, xie2024automating}). The resulting agent qualification improvements are reported in the second row of each setting in Tab. \ref{tab:combined_sq_summary}. Notably,  in the income and law school settings, where all features are causal and $f=h$, the qualification improvements are identical to the score increases. In the other three settings, qualification improvements are often substantially lower than score increases. This outcome is expected as non-causal features, while easier to modify, do not contribute to genuine improvements in qualification. \textbf{Nonetheless, we observe that LLMs achieve qualification improvements comparable to those of the theoretical model:} In all $5$ settings, the resulting improvements of theoretical model are similar to LLMs, while GPT-5 even outperforms the theoretical model in $2$ settings. These results suggest that LLMs generally provide effective strategies for helping agents improve their true qualifications. We also visualize the resulting qualification distributions. The plot for the law school setting is shown in Figure~\ref{fig:q_law_d1}, and the remaining plots are provided in the Appendix \ref{app:add_exp} (Fig.~\ref{fig:q_hiring_d1} through \ref{fig:q_pap_d1}).

\paragraph{Decision scenario 2.} In this scenario, the decision-maker deploys a logistic classifier $f$ while the ground-truth labeling function $h$ is a multilayer perceptron (MLP). This setup reflects practical constraints where decision-makers favor simpler, more interpretable models due to transparency requirements and lack of access to the true labeling function. Here, agent qualification improvement is evaluated using the more complex $h$, and the results are reported in the third row of each setting in Tab. \ref{tab:combined_sq_summary}. The findings are broadly consistent with those observed in decision scenario 1. That is, LLMs achieve qualification improvements comparable to those of the theoretical model, and GPT-5 outperforms the theoretical model in $3$ of the $5$ settings. We also visualize the resulting qualification distributions. The plot for the law school setting is shown in Fig.~\ref{fig:q_law_d2}, and the plots for the remaining settings are included in the Appendix \ref{app:add_exp} (Fig.~\ref{fig:q_hiring_d2} through \ref{fig:q_public_d2}).

\begin{figure*}[h]
    \centering
    \begin{subfigure}{0.32\textwidth}
        \centering
        \includegraphics[width= \linewidth]{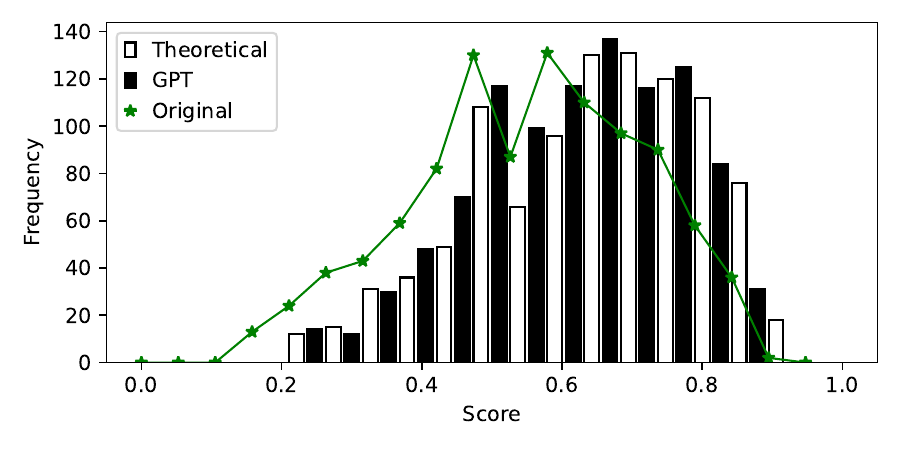}
        \caption{GPT-4o}
        \label{subfig:q_law_d2_4o}
    \end{subfigure}
    \hspace{0.1cm}
    \begin{subfigure}{0.32\textwidth}
        \centering
        \includegraphics[width= \linewidth]{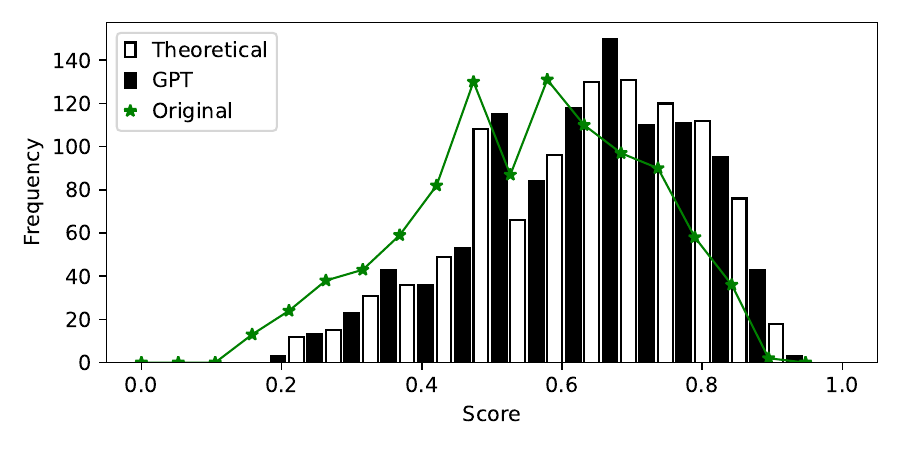}
        \caption{GPT-4.1}
        \label{subfig:q_law_d2_41}
    \end{subfigure}
    \hspace{0.1cm}
    \begin{subfigure}{0.32\textwidth}
        \centering
        \includegraphics[width= \linewidth]{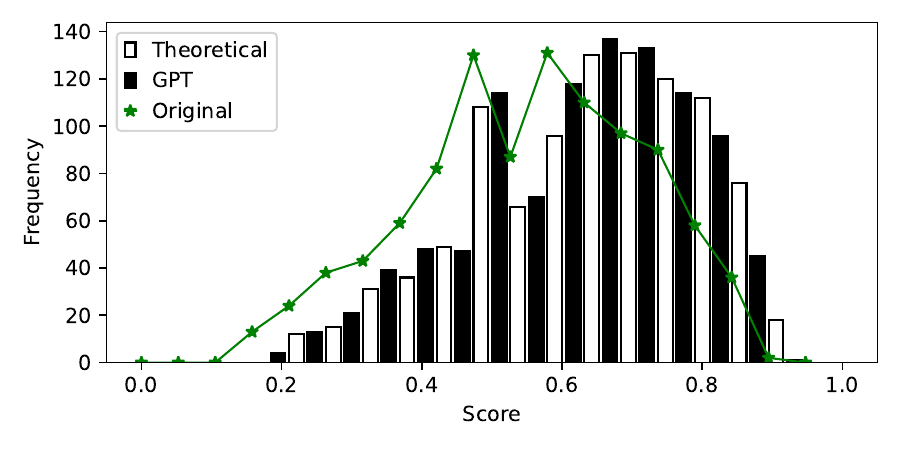}
        \caption{GPT-5}
        \label{subfig:q_law_d2_5}
    \end{subfigure}
    \caption{Comparison of the qualification distribution before and after agent responses in the \textbf{law school} setting for decision scenario 2. While the theoretical model yields the largest qualification improvement (evidenced by the rightward shift of the hollow bars), the LLMs (solid bars) also result in a considerable improvement, albeit slightly smaller in magnitude.}
    \label{fig:q_law_d2}
\end{figure*}

\begin{table*}[h!]
\centering
\caption{Unfairness (measured by score increase disparity and qualification improvement disparity) of the effort allocation strategies produced by LLMs and the theoretical model across different settings under decision scenario 1 (S1) and decision scenario 2 (S2). In each of the $5$ settings, LLMs achieve similar levels of unfairness to the theoretical model.}
\label{tab:combined_fair_summary}
\resizebox{0.85\textwidth}{!}{
\begin{tabular}{llcccc}
\toprule
\textbf{Setting} & \textbf{Metric}                                & \textbf{Theoretical} & \textbf{GPT-4o} & \textbf{GPT-4.1} & \textbf{GPT-5} \\
\midrule
\multirow{3}{*}{\textbf{Hiring}} 
    & Score increase disparity                      & $0.015$            & $0.078$          & $0.016$    &$0.051$      \\
    & Qualification improvement disparity (S1)      & $0.014$            & $0.078$          & $0.016$    &$0.047$       \\
    & Qualification improvement disparity (S2)      & $-0.001$            & $0.068$         & $0.009$    &$0.030$       \\
\midrule
\multirow{3}{*}{\textbf{Income}} 
    & Score increase disparity                      & $0.014$            & $0.008$          & $0.008$     &$0.004$     \\
    & Qualification improvement disparity (S1)      & $0.013$            & $0.008$          & $0.008$     &$0.004$      \\
    & Qualification improvement disparity (S2)      & $0.007$            & $0.006$          & $0.006$     &$0.004$     \\
\midrule
\multirow{3}{*}{\textbf{Law school}} 
    & Score increase disparity                      & $0.003$            & $0.000$          & $0.000$    & $0.002$      \\
    & Qualification improvement disparity (S1)      & $0.003$            & $0.000$          & $0.000$    & $0.002$     \\
    & Qualification improvement disparity (S2)      & $0.012$            & $0.009$          & $0.010$    & $0.011$      \\
\midrule
\multirow{3}{*}{\textbf{Loan approval}} 
    & Score increase disparity                      & $-0.012$           & $-0.007$         & $-0.005$  & $0.002$       \\
    & Qualification improvement disparity (S1)      & $-0.001$           & $0.003$          & $0.003$    & $0.005$       \\
    & Qualification improvement disparity (S2)      & $0.003$            & $0.007$          & $0.02$     & $0.05$     \\
\midrule
\multirow{3}{*}{\textbf{Public assistance program}} 
    & Score increase disparity                      & $-0.002$           & $0.00$          & $-0.003$   & $0.03$       \\
    & Qualification improvement disparity (S1)      & $-0.004$           & $-0.004$          & $-0.007$   & $0.03$       \\
    & Qualification improvement disparity (S2)      & $0.104$            & $-0.145$         & $0.140$    & $0.129$      \\
\bottomrule
\end{tabular}
}
\end{table*}

\subsection{Unfairness of the Effort Allocation Strategies}\label{subsec:fairness}

In this section, we examine whether LLM-generated strategies lead to disparate outcomes in score increases or qualification improvements across different social groups. Specifically, we define the unfairness of an effort allocation strategy as the disparity in score increases or qualification improvements between social groups, assuming all agents follow the suggested strategy. This notion of unfairness is conceptually aligned with fairness metrics such as \textit{equal improvability} \cite{guldogan2022equal} and \textit{bounded effort} \cite{heidari2019long}, which measure fairness by assessing whether different groups achieve similar improvements after best responding to a decision policy. However, rather than evaluating the fairness of the decision policy itself, our focus is on the fairness of the effort allocation strategies.

Tab. \ref{tab:combined_fair_summary} reports the unfairness measured in terms of score increases and qualification improvements across the two decision scenarios in each of the five settings. The results show that, overall, the effort allocation strategies generated by LLMs achieve fairness levels closely aligned with those of the theoretical model. \textbf{Notably, in each of the 15 comparisons, there exists at least one LLM whose unfairness measure lies within 0.03 of the corresponding theoretical value.} These consistently small gaps provide no evidence that LLM-generated strategies exacerbate unfairness relative to the theoretical benchmark.

\section{Conclusions \& Societal Impacts}
This work provides a comprehensive comparison between the theoretical best response model in strategic classification and LLM-generated strategic responses. We consider practical scenarios in which individuals with limited knowledge of the decision policy act based on suggestions from LLMs, and we show that commercial LLMs can effectively guide human agents in strategically responding to policy. Notably, the outcomes of LLM achieve levels of qualification improvement and fairness comparable to those of the theoretical model. Furthermore, LLMs tend to produce diverse effort allocation plans tailored to individual agents, often promoting more balanced distributions of effort across settings. However, as our findings are based on only five simulation settings, future work is needed to more comprehensively assess the fairness implications of LLM-generated strategies.

\bibliographystyle{ACM-Reference-Format}
\bibliography{sample-base}

\appendix

\newpage

\section{All prompts}\label{app:prompt}

Prompts for hiring setting:

\small
\begin{lstlisting}
You are a career advisor guiding users to change their personal profiles to be hired. Users will provide their current career profile as a dictionary of features and their values. Based on these features, Your task is to recommend the optimal effort allocation strategies that will improve the probability of the user getting the job.

### Requirement of your recommendation strategies

1. Your recommendation strategy must be based on the unique user's provided features and your knowledge and reasoning to help them get the job.

1. Your recommendation strategy must be a JSON dictionary containing up to five strategies for
affecting their features "RevolvingUtilizationOfUnsecuredLines", "education", "YearsCode", "PreviousSalary", and "ComputerSkills".

2. Strategies include a Direction ("increase" or "decrease"), and Effort (the amount of effort going to changing that feature in the given Direction). Do not use any direction other than increase or decrease.

3. Effort is valid as long as it is a non-negative number. Although there is no effort budeget, for each unit of effort, the user will pay a cost of the square of this effort divided by 2 (e.g., 0.5 effort will incur 0.5^2/2 = 0.125 cost). While the reward of the user will be the amount of probability improvement (maximum reward is 1 since the largest possible probability is 1) to get the job after changing their profiles following your strategy. You must consider whether the cost is worthwhile compared to the reward.

### Mandatory output schema

Your output must have the following JSON schema **without** any additional explanation:

{
    "education": {
        "Direction": "increase" or "decrease" or "N/A" if "Effort" is 0,
        "Effort": "the amount of effort allocated to this feature"
    },
    "YearsCode": {
        "Direction": "increase" or "decrease" or "N/A" if "Effort" is 0,
        "Effort": "the amount of effort allocated to this feature"
    },
    "PreviousSalary": {
        "Direction": "increase" or "decrease" or "N/A" if "Effort" is 0,
        "Effort": "the amount of effort allocated to this feature"
    },
    "ComputerSkills": {
        "Direction": "increase" or "decrease" or "N/A" if "Effort" is 0,
        "Effort": "the amount of effort allocated to this feature"
    },
}

Note that you are allowed to allocate 0 effort to some feature. But when effort is 0, the "Direction" must be "N/A".
\end{lstlisting}
\normalsize

Prompts for income setting:

\small
\begin{lstlisting}
    You are a financial advisor guiding users to change their personal profiles to increase their income. Users will provide their current personal profiles as a dictionary of features and their values. Based on these features, Your task is to recommend the optimal effort allocation strategies that will improve the probability of the user getting a high income.

### Requirement of your recommendation strategies

1. Your recommendation strategy must be based on the unique user's provided features and your knowledge and reasoning to help them increase the income.

1. Your recommendation strategy must be a JSON dictionary containing up to five strategies for
affecting their features "SCHL" (education level), "WKHP" (working hours per week).

2. Strategies include a Direction ("increase" or "decrease"), and Effort (the amount of effort going to changing that feature in the given Direction). Do not use any direction other than increase or decrease.

3. Effort is valid as long as it is a non-negative number. Although there is no effort budeget, for each unit of effort, the user will pay a cost of the square of this effort divided by 2 (e.g., 0.5 effort will incur 0.5^2/2 = 0.125 cost). While the reward of the user will be the amount of probability improvement (maximum reward is 1 since the largest possible probability is 1) to get a high income after changing their profiles following your strategy. You must consider whether the cost is worthwhile compared to the reward.

### Mandatory output schema

Your output must have the following JSON schema **without** any additional explanation:

{
    "SCHL": {
        "Direction": "increase" or "decrease" or "N/A" if "Effort" is 0,
        "Effort": "the amount of effort allocated to this feature"
    },
    "WKHP": {
        "Direction": "increase" or "decrease" or "N/A" if "Effort" is 0,
        "Effort": "the amount of effort allocated to this feature"
    },
}

Note that you are allowed to allocate 0 effort to some feature. But when effort is 0, the "Direction" must be "N/A".
\end{lstlisting}
\normalsize

Prompts for loan approval setting:

\small
\begin{lstlisting}
You are a financial advisor guiding users to change their financial profiles to get loan approval. Users will provide their current financial profiles as a dictionary of features and their values. Based on these features, Your task is to recommend the optimal effort allocation strategies that will improve the probability of the user getting a loan approval. 

### Requirement of your recommendation strategies

1. Your recommendation strategy must be based on the unique user's provided features and your knowledge and reasoning to help them get the loan approval.

1. Your recommendation strategy must be a JSON dictionary containing up to five strategies for
affecting their features "RevolvingUtilizationOfUnsecuredLines", "NumberOfOpenCreditLinesAndLoans", "Number
RealEstateLoansOrLines", "MonthlyIncome", and "DebtRatio".

2. Strategies include a Direction ("increase" or "decrease"), and Effort (the amount of effort going to changing that feature in the given Direction). Do not use any direction other than increase or decrease.

3. Effort is valid as long as it is a non-negative number. Although there is no effort budeget, for each unit of effort, the user will pay a cost of the square of this effort divided by 2 (e.g., 0.5 effort will incur 0.5^2/2 = 0.125 cost). While the reward of the user will be the amount of probability improvement (maximum reward is 1 since the largest possible probability is 1) to get loan approval after changing their profiles following your strategy. You must consider whether the cost is worthwhile compared to the reward.

### Mandatory output schema

Your output must have the following JSON schema **without** any additional explanation:

{
    "RevolvingUtilizationOfUnsecuredLines": {
        "Direction": "increase" or "decrease" or "N/A" if "Effort" is 0,
        "Effort": "the amount of effort allocated to this feature"
    },
    "NumberOfOpenCreditLinesAndLoans": {
        "Direction": "increase" or "decrease" or "N/A" if "Effort" is 0,
        "Effort": "the amount of effort allocated to this feature"
    },
    "NumberRealEstateLoansOrLines": {
        "Direction": "increase" or "decrease" or "N/A" if "Effort" is 0,
        "Effort": "the amount of effort allocated to this feature"
    },
    "MonthlyIncome": {
        "Direction": "increase" or "decrease" or "N/A" if "Effort" is 0,
        "Effort": "the amount of effort allocated to this feature"
    },
    "DebtRatio": {
        "Direction": "increase" or "decrease" or "N/A" if "Effort" is 0,
        "Effort": "the amount of effort allocated to this feature"
    },  
}

Note that you are allowed to allocate 0 effort to some feature. But when effort is 0, the "Direction" must be "N/A".
\end{lstlisting}
\normalsize

Prompts for law school setting:

\begin{lstlisting}
You are a education advisor guiding users to change their personal profiles to be admitted by law schools. Users will provide their current academic profiles as a dictionary of features and their values. Based on these features, Your task is to recommend the optimal effort allocation strategies that will improve the probability of the user getting admitted.

### Requirement of your recommendation strategies

1. Your recommendation strategy must be based on the unique user's provided features and your knowledge and reasoning to help them get admitted.

1. Your recommendation strategy must be a JSON dictionary containing up to two strategies for
affecting their features "UGPA" (undergrad GPA), "LSAT" (LSAT score).

2. Strategies include a Direction ("increase" or "decrease"), and Effort (the amount of effort going to changing that feature in the given Direction). Do not use any direction other than increase or decrease.

3. Effort is valid as long as it is a non-negative number. Although there is no effort budeget, for each unit of effort, the user will pay a cost of the square of this effort divided by 2 (e.g., 0.5 effort will incur 0.5^2/2 = 0.125 cost). While the reward of the user will be the amount of probability improvement (maximum reward is 1 since the largest possible probability is 1) to get admitted after changing their profiles following your strategy. You must consider whether the cost is worthwhile compared to the reward.

### Mandatory output schema

Your output must have the following JSON schema **without** any additional explanation:

{
    "UGPA": {
        "Direction": "increase" or "decrease" or "N/A" if "Effort" is 0,
        "Effort": "the amount of effort allocated to this feature"
    },
    "LSAT": {
        "Direction": "increase" or "decrease" or "N/A" if "Effort" is 0,
        "Effort": "the amount of effort allocated to this feature"
    },
}

Note that you are allowed to allocate 0 effort to some feature. But when effort is 0, the "Direction" must be "N/A".
\end{lstlisting}

\section{The Connection Between Def. \ref{def:sc_game} and Def. \ref{def:br}}\label{app:proof}

In this section, we prove that when the decision policy is linear, i.e., $f = \beta^Tx + \beta_0$ and the cost function is quadratic, i.e., $c(x,z) = (z-x)^TB(z-x)$ where $B$ is symmetric and semidefinite (i.e., moving will never incur negative costs), then for each $c$, there exists $w, \delta$ to let the agent best response defined in Def. \ref{def:br} be equivalent to Def. \ref{def:sc_game}.

\begin{proof}
    We firstly work out the best response using Def. \ref{def:sc_game}, where we need to work out:
    \begin{align*}
        \arg \max_{z \in \mathcal{X}} f(z) - c(x,z) = \beta^Tz - (z-x)^TB(z-x)
    \end{align*}
    Take the derivative with respect to $x$ of the above equation to get $\beta^T - 2 \cdot B(z-x) = 0$. Thus, $z = x + \frac{\beta^T B^{-1}}{2}$.
    Next, consider the best response using Def. \ref{def:br}, where we need to work out:
    \begin{align*}
        \arg \max_{e \in {\mathbb{R}^{+}}^d} B(x(e)) = \beta^Tx + \beta^TW e - \frac{1}{2}e^Te
    \end{align*}
    then $e = W\beta$ and $z = x + W^2\beta$. Since $B$ is semidefinite, there must exist $W$ to satisfy $W^2 = B^{-1}$.
\end{proof}

\section{Additional Experimental Results}\label{app:add_exp}

\paragraph{The effort allocation for GPT-4o and GPT-4.1.}

\begin{figure*}[h]
    \centering
    \begin{subfigure}{0.43\textwidth}
        \centering
        \includegraphics[width= \linewidth]{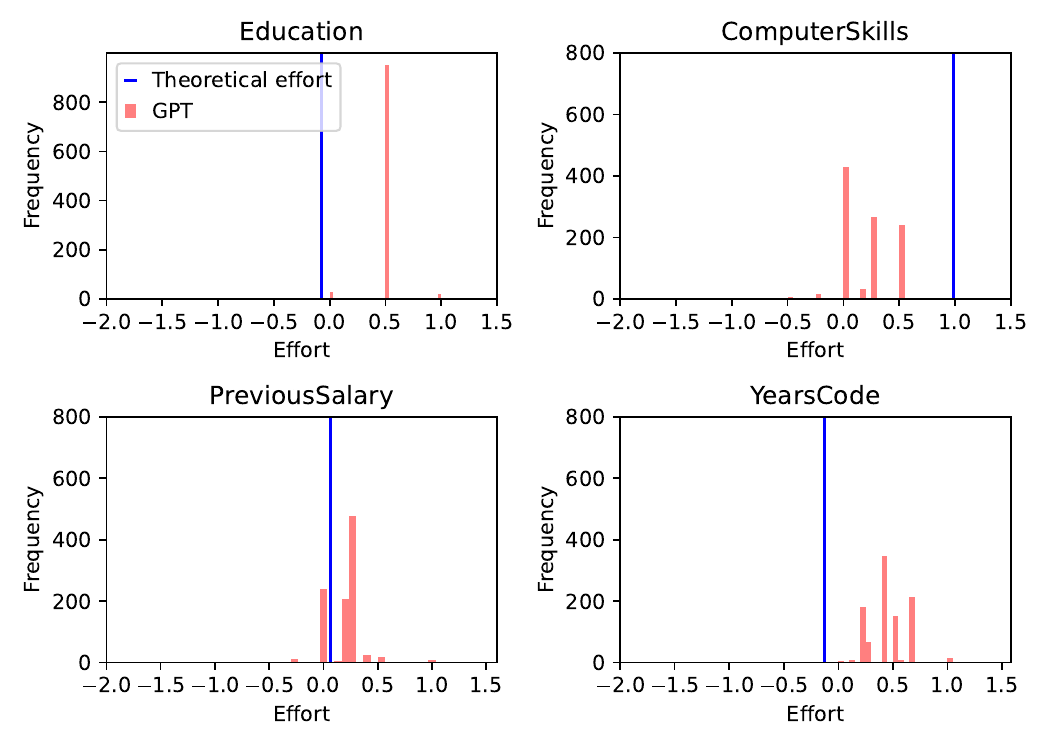}
        \caption{GPT-4o}
        \label{subfig:effort_hiring_d12_4o}
    \end{subfigure}
    \hspace{0.1cm}
    \begin{subfigure}{0.43\textwidth}
        \centering
        \includegraphics[width= \linewidth]{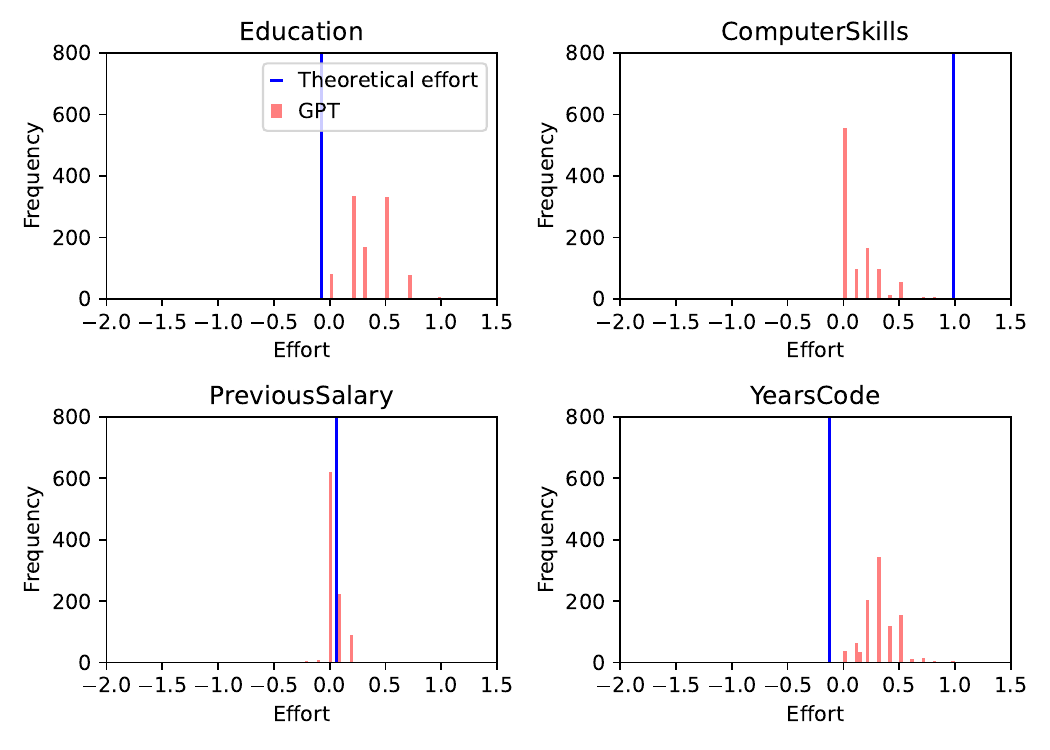}
        \caption{GPT-4.1}
        \label{subfig:effort_hiring_d12_41}
    \end{subfigure}
    \caption{Comparison of effort allocations between the theoretical model and LLM in the \textbf{hiring} setting, produced by GPT-4o and GPT-4.1. As the decision policies are identical in both decision scenarios, the resulting effort allocations are also the same.}
    \label{fig:effort_hiring_d12}
\end{figure*}

\begin{figure*}[h]
    \centering
    \begin{subfigure}{0.47\textwidth}
        \centering
        \includegraphics[width= \linewidth]{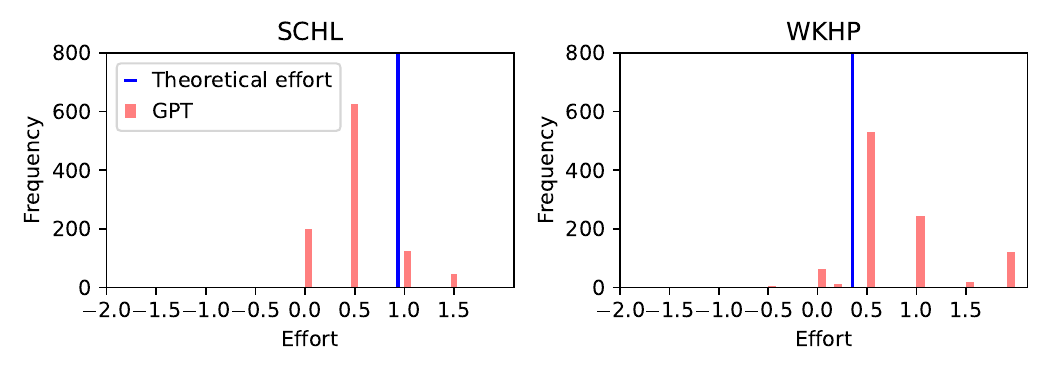}
        \caption{GPT-4o}
        \label{subfig:effort_income_d12_4o}
    \end{subfigure}
    \hspace{0.1cm}
    \begin{subfigure}{0.47\textwidth}
        \centering
        \includegraphics[width= \linewidth]{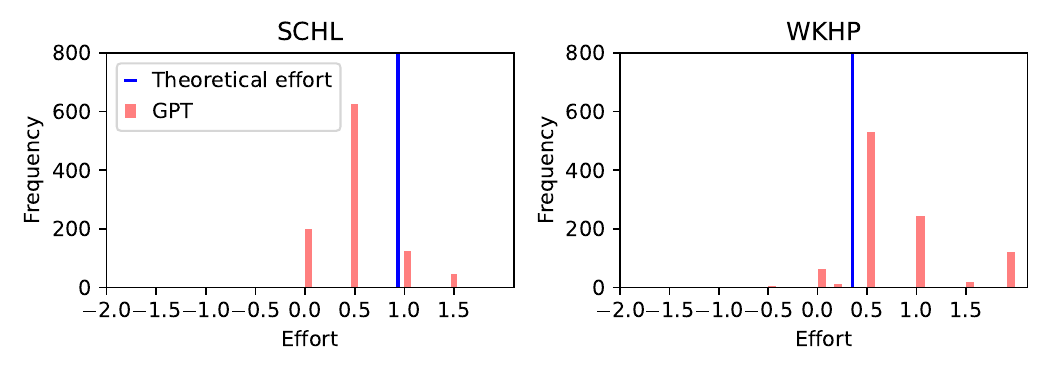}
        \caption{GPT-4.1}
        \label{subfig:effort_income_d12_41}
    \end{subfigure}
    \caption{Comparison of effort allocations between the theoretical model and LLM in the \textbf{income} setting, produced by GPT-4o and GPT-4.1. As the decision policies are identical in both decision scenarios, the resulting effort allocations are also the same.}
    \label{fig:effort_income_d12}
\end{figure*}

\begin{figure*}[h]
    \centering
    \begin{subfigure}{0.47\textwidth}
        \centering
        \includegraphics[width= \linewidth]{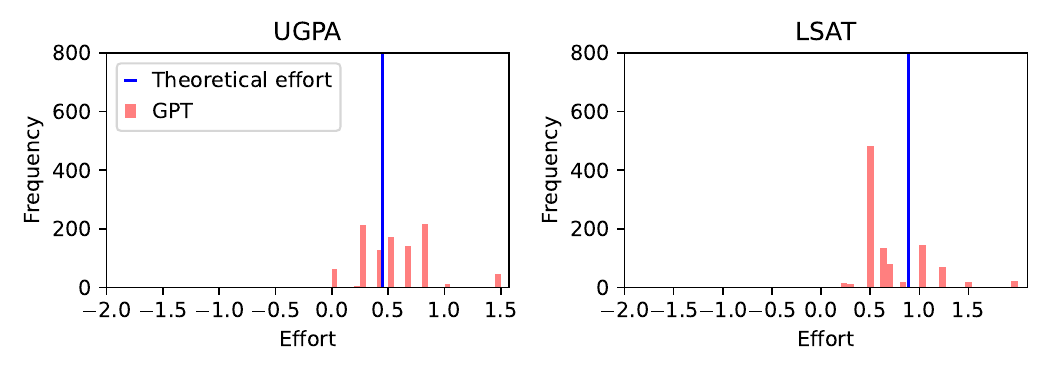}
        \caption{GPT-4o}
        \label{subfig:effort_law_d12_4o}
    \end{subfigure}
    \hspace{0.1cm}
    \begin{subfigure}{0.47\textwidth}
        \centering
        \includegraphics[width= \linewidth]{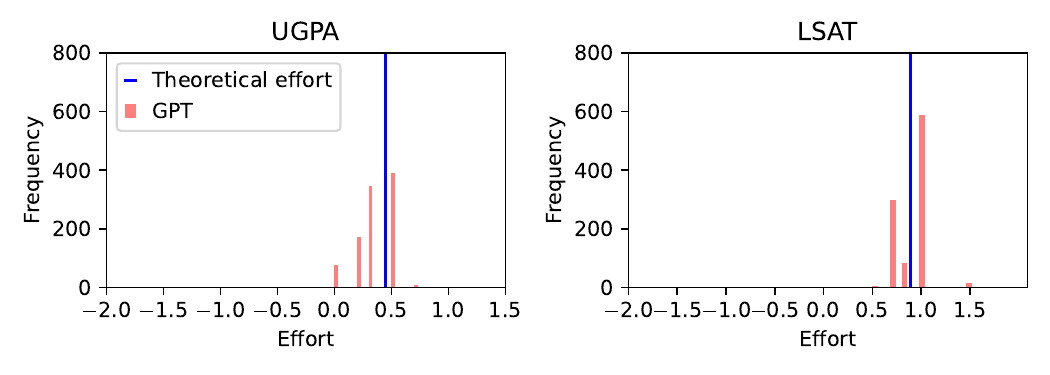}
        \caption{GPT-4.1}
        \label{subfig:effort_law_d12_41}
    \end{subfigure}
    \caption{Comparison of effort allocations between the theoretical model and LLM in the \textbf{law school} setting, produced by GPT-4o and GPT-4.1. As the decision policies are identical in both decision scenarios, the effort allocations are also the same.}
    \label{fig:effort_law_d12}
\end{figure*}

\begin{figure*}[h]
    \centering
    \begin{subfigure}{0.76\textwidth}
        \centering
        \includegraphics[width= \linewidth]{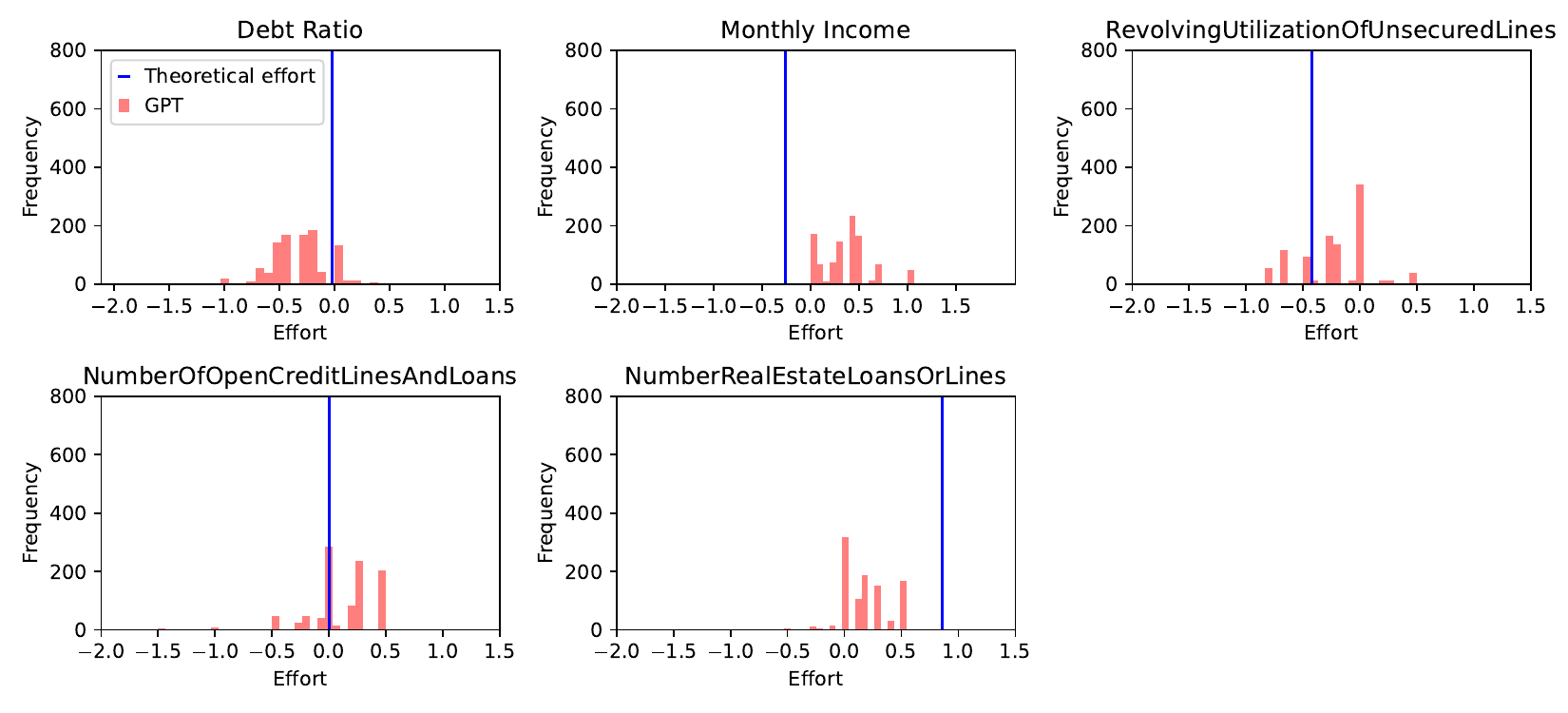}
        \caption{GPT-4o}
        \label{subfig:effort_credit_d12_4o}
    \end{subfigure}
    \hfill
    \begin{subfigure}{0.76\textwidth}
        \centering
        \includegraphics[width= \linewidth]{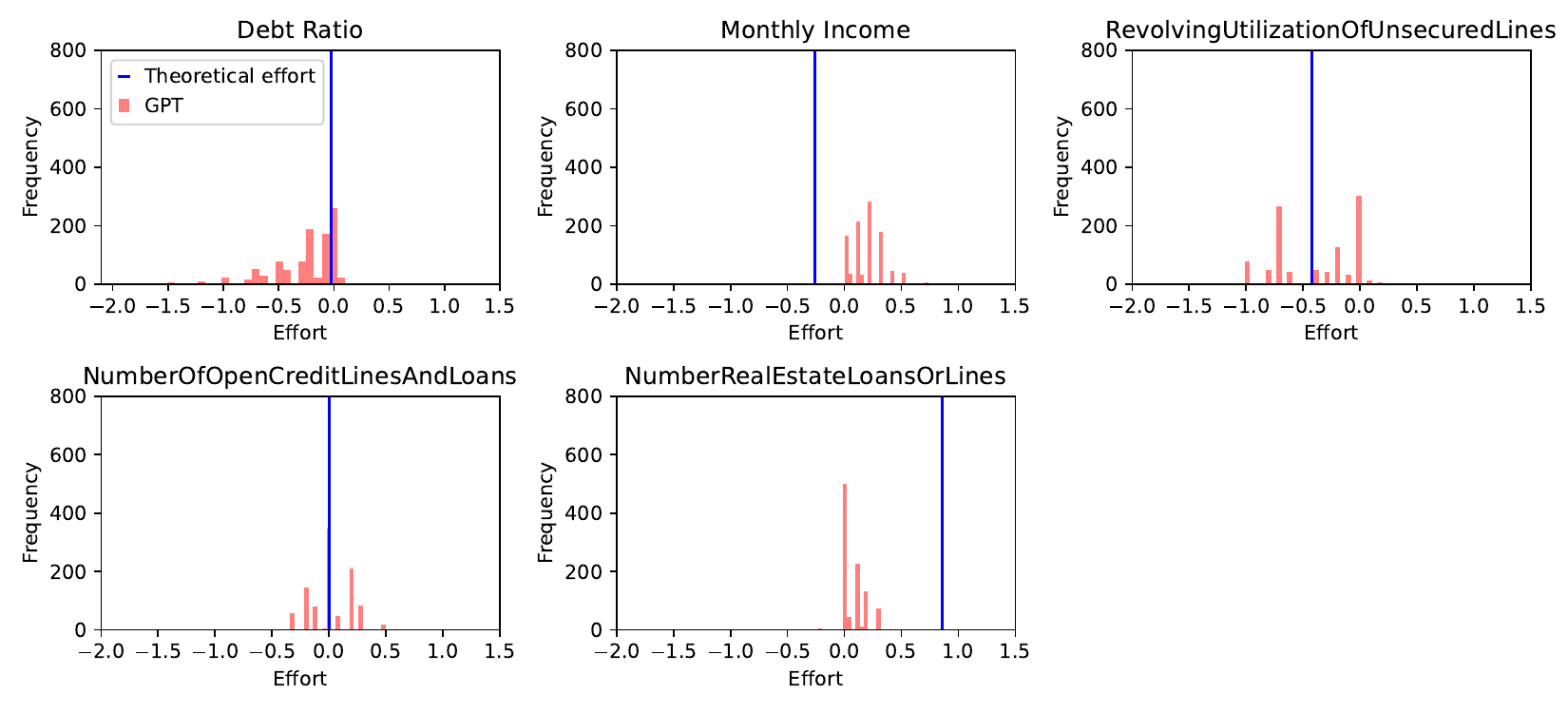}
        \caption{GPT-4.1}
        \label{subfig:effort_credit_d12_41}
    \end{subfigure}
    \caption{Comparison of effort allocations between the theoretical model and LLM in the \textbf{loan approval} setting produced by GPT-4o and GPT-4.1. As the decision policies are identical in both decision scenarios, the effort allocations are also the same. }
    \label{fig:effort_credit_d12}
\end{figure*}

\begin{figure*}[h]
    \centering
    \begin{subfigure}{0.75\textwidth}
        \centering
        \includegraphics[width= \linewidth]{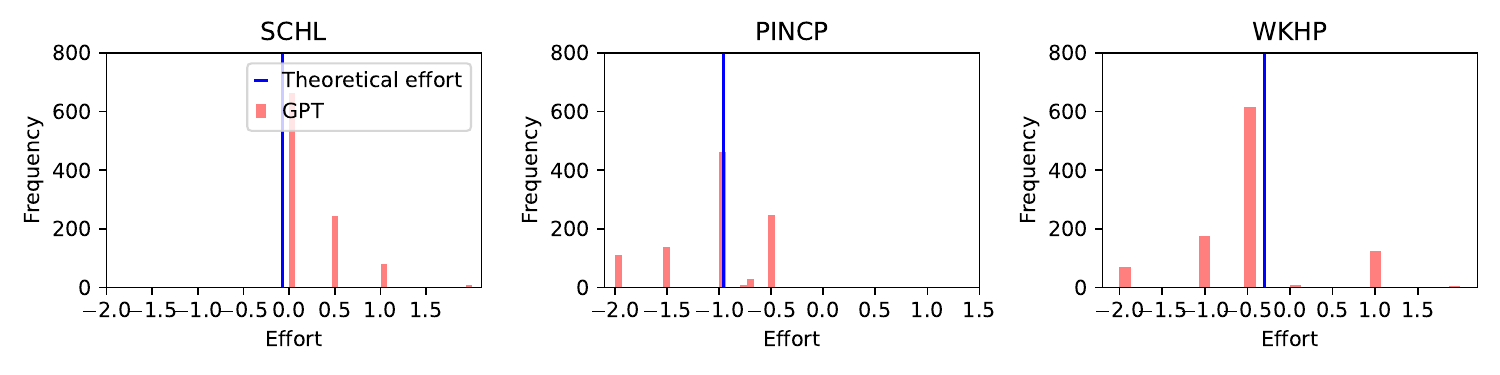}
        \caption{GPT-4o}
        \label{subfig:effort_pap_d12_4o}
    \end{subfigure}
    \hfill
    \begin{subfigure}{0.75\textwidth}
        \centering
        \includegraphics[width= \linewidth]{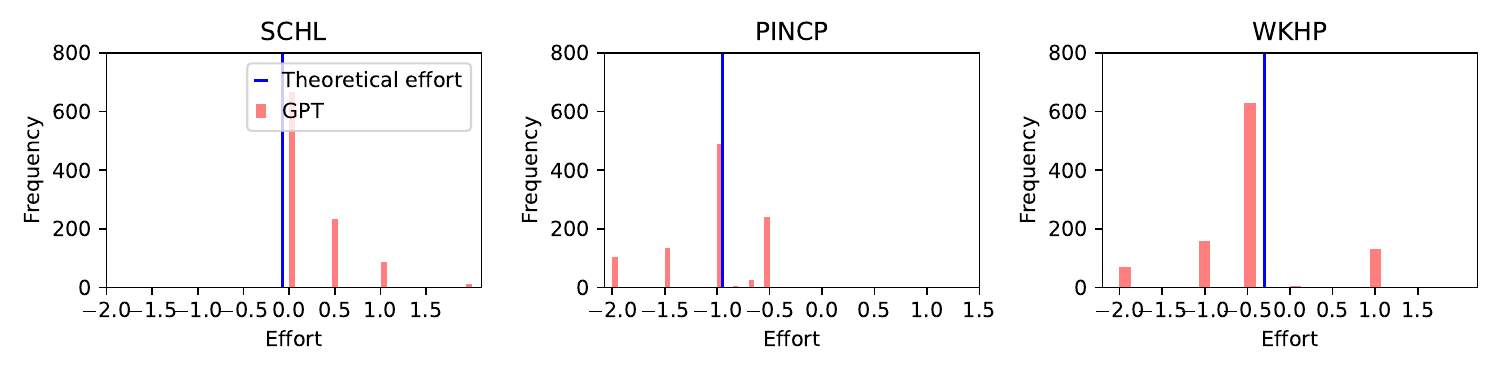}
        \caption{GPT-4.1}
        \label{subfig:effort_pap_d12_41}
    \end{subfigure}
    \caption{Comparison of effort allocations between the theoretical model and LLM in the \textbf{public assistance} setting produced by GPT-4o and GPT-4.1. As the policies are identical in both decision scenarios, the effort allocations are also the same.}
    \label{fig:effort_pap_d12}
\end{figure*}

\paragraph{Additional plots of score/qualification distribution changes.}

\begin{figure*}[h]
    \centering
    \begin{subfigure}{0.32\textwidth}
        \centering
        \includegraphics[width= \linewidth]{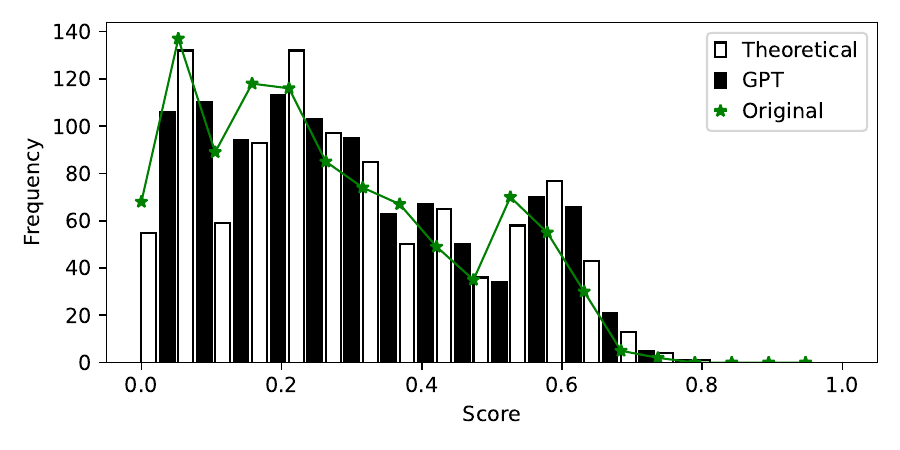} 
        \caption{GPT-4o}
        \label{subfig:score_income_d1_4o}
    \end{subfigure}
    \hfill
    \begin{subfigure}{0.32\textwidth}
        \centering
        \includegraphics[width= \linewidth]{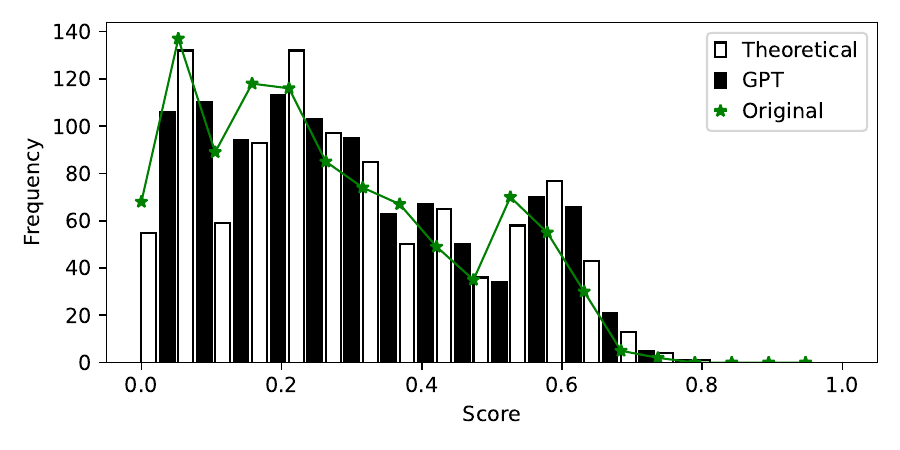}
        \caption{GPT-4.1}
        \label{subfig:score_income_d1_41}
    \end{subfigure}
    \hfill
    \begin{subfigure}{0.32\textwidth}
        \centering
        \includegraphics[width= \linewidth]{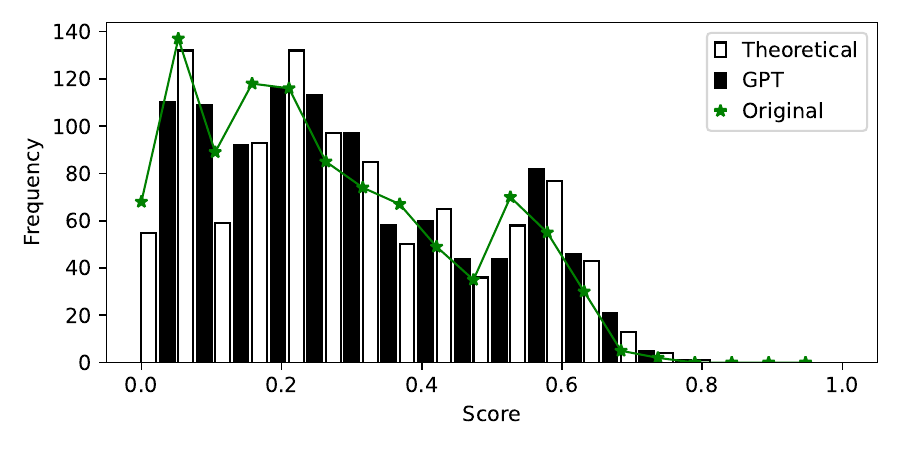}
        \caption{GPT-5}
        \label{subfig:score_income_d1_5}
    \end{subfigure}
    \caption{Comparison of the score distribution before/after agent response in the income setting.}
    \label{fig:score_income_d1}
\end{figure*}

\begin{figure*}[h]
    \centering
    \begin{subfigure}{0.32\textwidth}
        \centering
        \includegraphics[width= \linewidth]{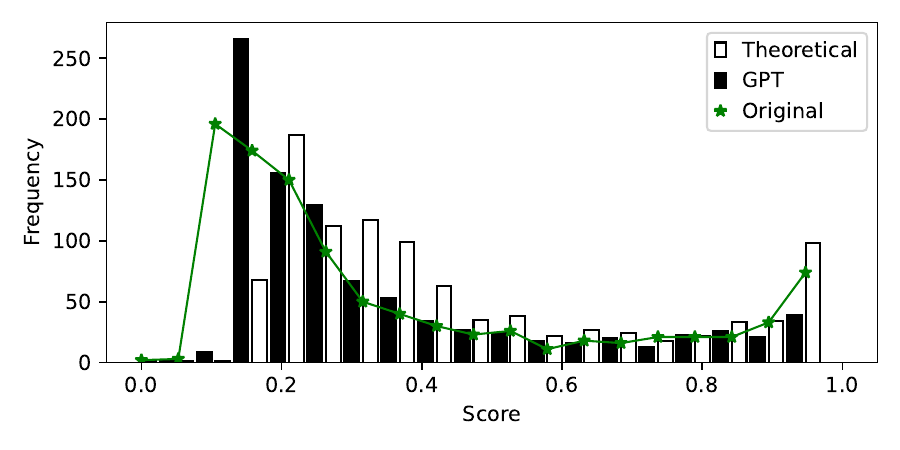} 
        \caption{GPT-4o}
        \label{subfig:score_credit_d1_4o}
    \end{subfigure}
    \hfill
    \begin{subfigure}{0.32\textwidth}
        \centering
        \includegraphics[width= \linewidth]{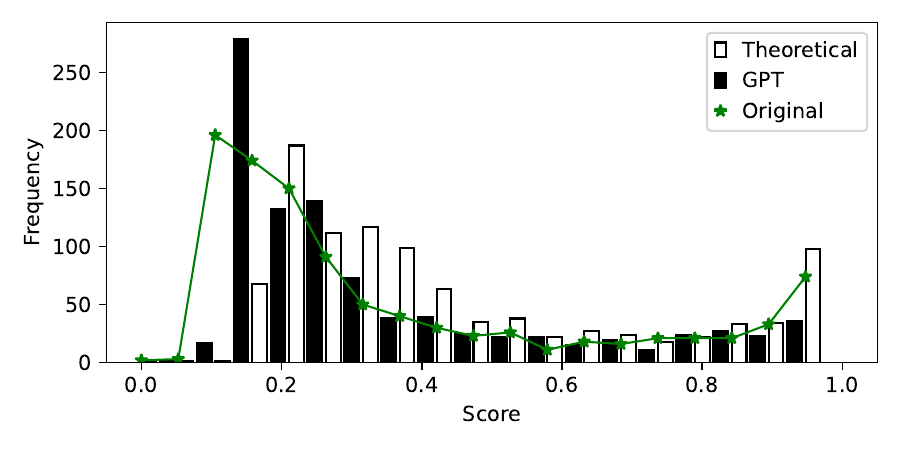}
        \caption{GPT-4.1}
        \label{subfig:score_credit_d1_41}
    \end{subfigure}
    \hfill
    \begin{subfigure}{0.32\textwidth}
        \centering
        \includegraphics[width= \linewidth]{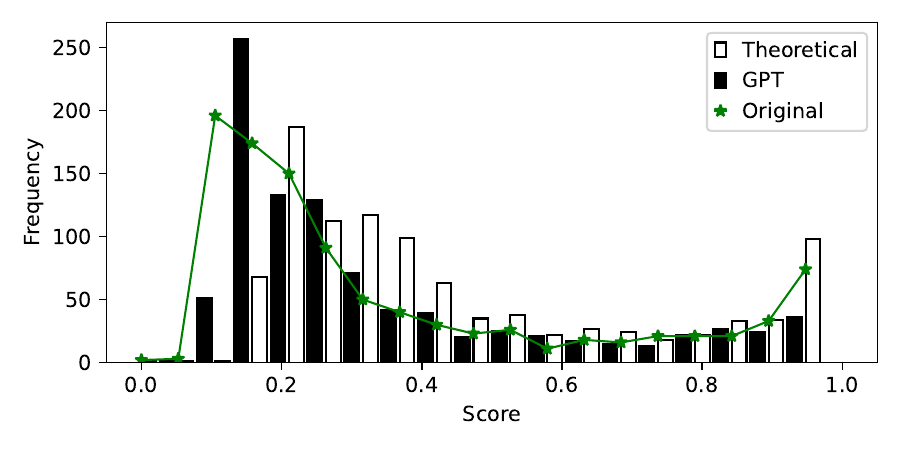}
        \caption{GPT-5}
        \label{subfig:score_credit_d1_5}
    \end{subfigure}
    \caption{Comparison of the score distribution before/after agent response in the loan approval setting.}
    \label{fig:score_credit_d1}
\end{figure*}

\begin{figure*}[h]
    \centering
    \begin{subfigure}{0.32\textwidth}
        \centering
        \includegraphics[width= \linewidth]{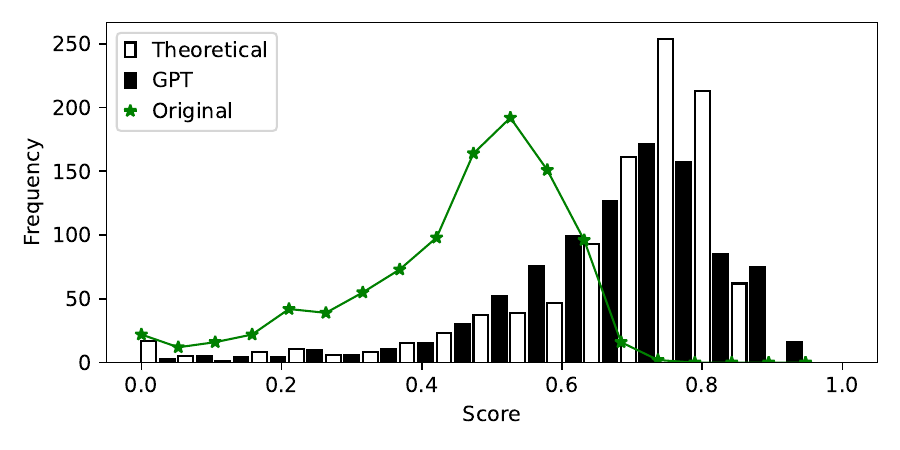} 
        \caption{GPT-4o}
        \label{subfig:score_pap_d1_4o}
    \end{subfigure}
    \hfill
    \begin{subfigure}{0.32\textwidth}
        \centering
        \includegraphics[width= \linewidth]{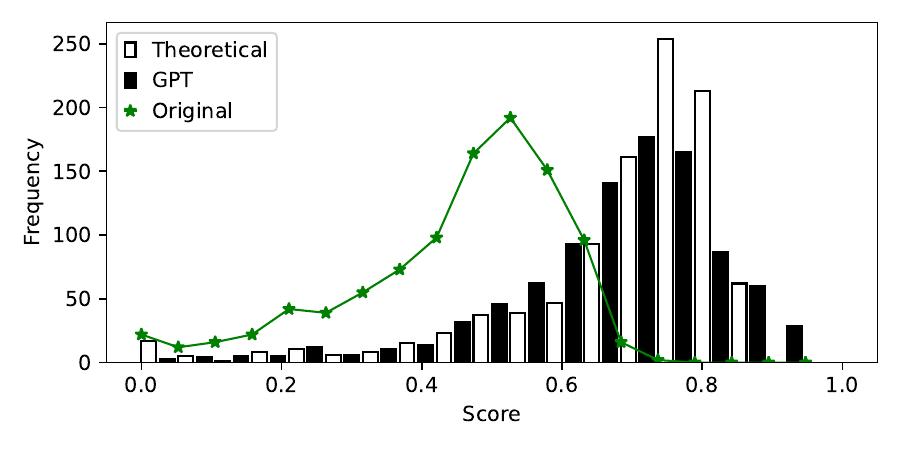}
        \caption{GPT-4.1}
        \label{subfig:score_pap_d1_41}
    \end{subfigure}
    \hfill
    \begin{subfigure}{0.32\textwidth}
        \centering
        \includegraphics[width= \linewidth]{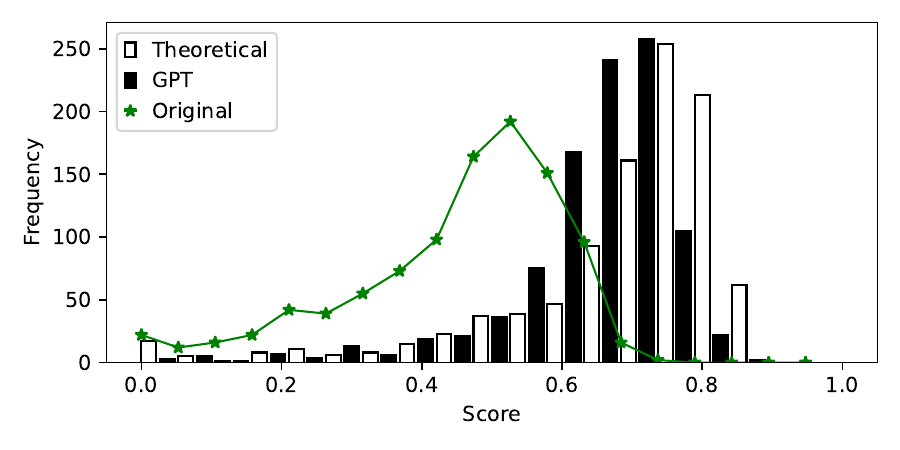}
        \caption{GPT-5}
        \label{subfig:score_pap_d1_5}
    \end{subfigure}
    \caption{Comparison of the score distribution before/after agent response in the public assistance program setting.}
    \label{fig:score_pap_d1}
\end{figure*}

\begin{figure*}[h]
    \centering
    \begin{subfigure}{0.32\textwidth}
        \centering
        \includegraphics[width= \linewidth]{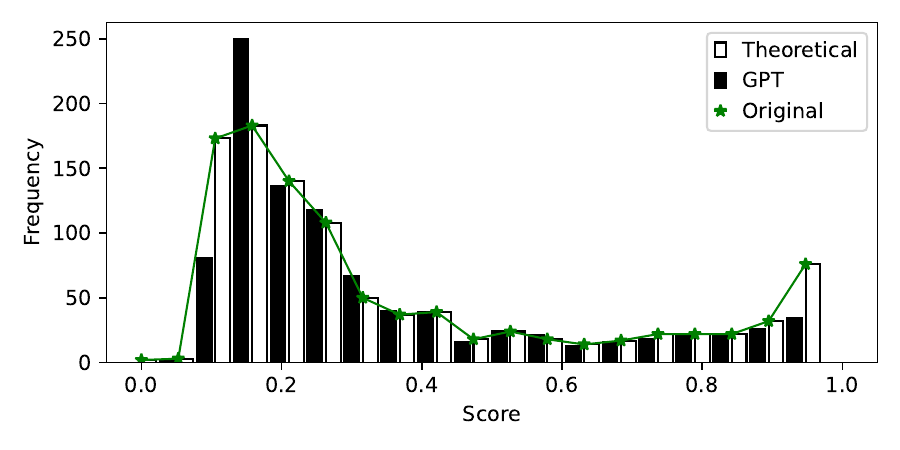}
        \caption{GPT-4o}
        \label{subfig:q_credit_d1_4o}
    \end{subfigure}
    \hfill
    \begin{subfigure}{0.32\textwidth}
        \centering
        \includegraphics[width= \linewidth]{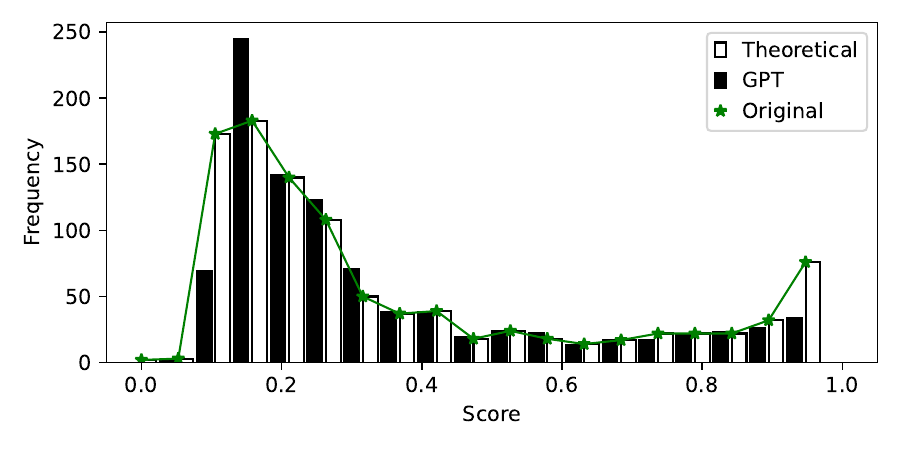}
        \caption{GPT-4.1}
        \label{subfig:q_credit_d1_41}
    \end{subfigure}
    \hfill
    \begin{subfigure}{0.32\textwidth}
        \centering
        \includegraphics[width= \linewidth]{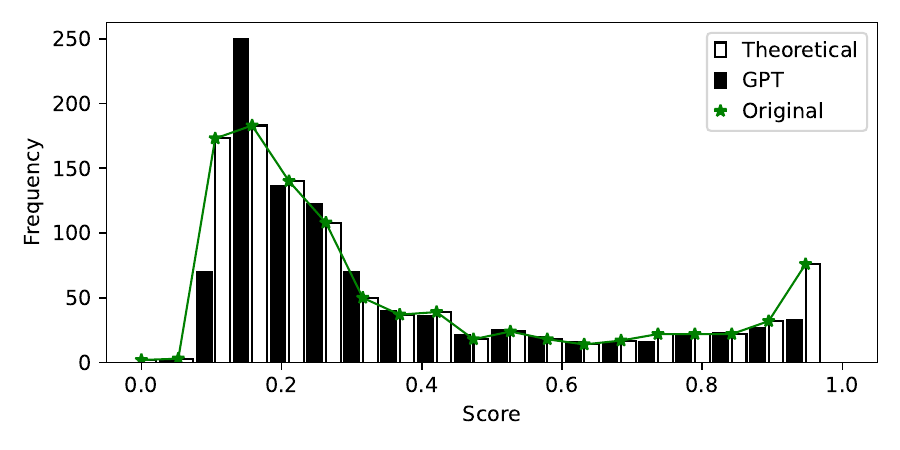}
        \caption{GPT-5}
        \label{subfig:q_credit_d1_5}
    \end{subfigure}
    \caption{Comparison of the qualification distribution before/after agent response in the loan approval setting and for decision scenario 1.}
    \label{fig:q_credit_d1}
\end{figure*}

\begin{figure*}[h]
    \centering
    \begin{subfigure}{0.32\textwidth}
        \centering
        \includegraphics[width= \linewidth]{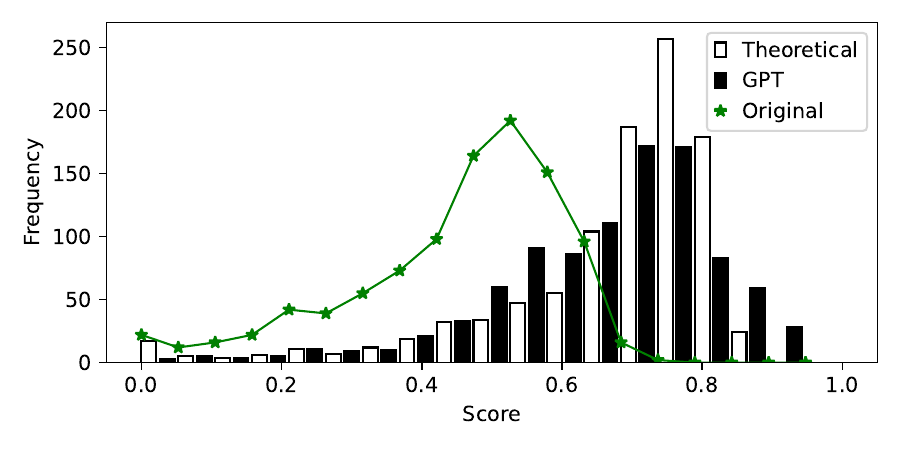}
        \caption{GPT-4o}
        \label{subfig:q_pap_d1_4o}
    \end{subfigure}
    \hfill
    \begin{subfigure}{0.32\textwidth}
        \centering
        \includegraphics[width= \linewidth]{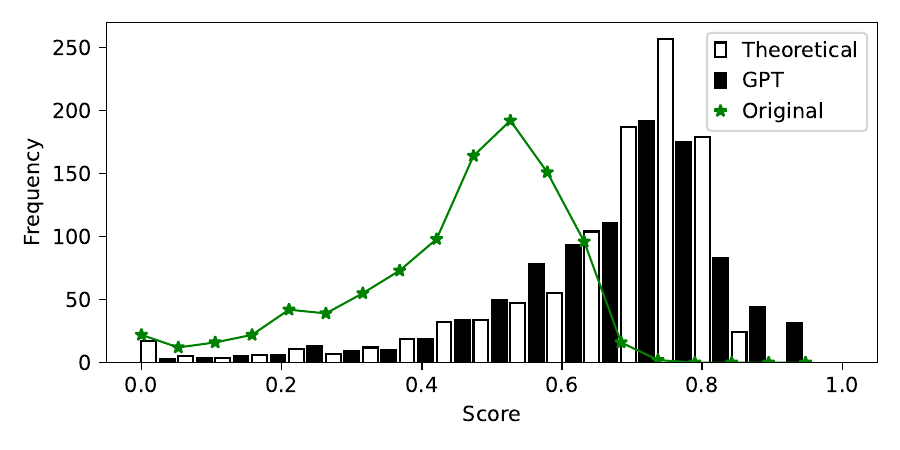}
        \caption{GPT-4.1}
        \label{subfig:q_pap_d1_41}
    \end{subfigure}
    \hfill
    \begin{subfigure}{0.32\textwidth}
        \centering
        \includegraphics[width= \linewidth]{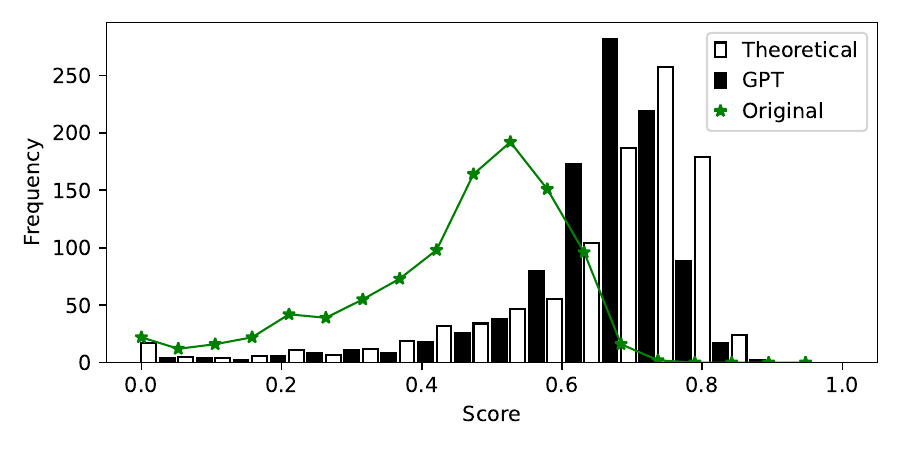}
        \caption{GPT-5}
        \label{subfig:q_pap_d1_5}
    \end{subfigure}
    \caption{Comparison of the qualification distribution before/after agent response in the public assistance program setting and for decision scenario 1.}
    \label{fig:q_pap_d1}
\end{figure*}

\begin{figure*}[h]
    \centering
    \begin{subfigure}{0.32\textwidth}
        \centering
        \includegraphics[width= \linewidth]{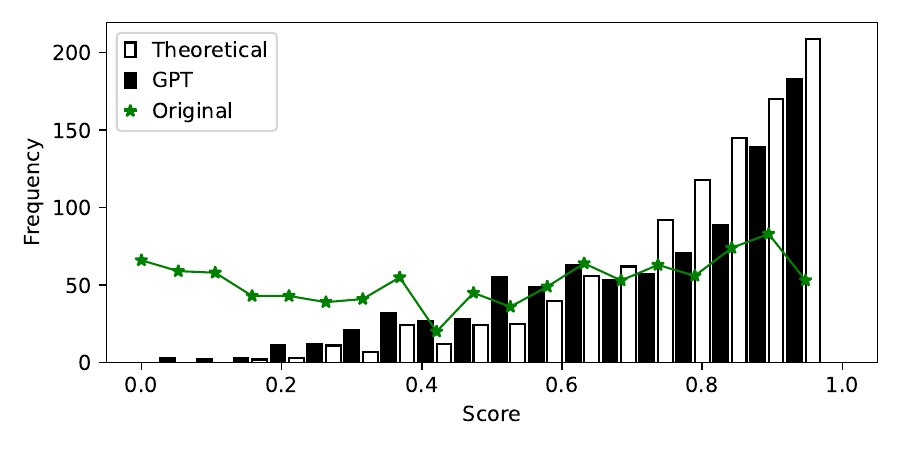}
        \caption{GPT-4o}
        \label{subfig:q_hiring_d2_4o}
    \end{subfigure}
    \hfill
    \begin{subfigure}{0.32\textwidth}
        \centering
        \includegraphics[width= \linewidth]{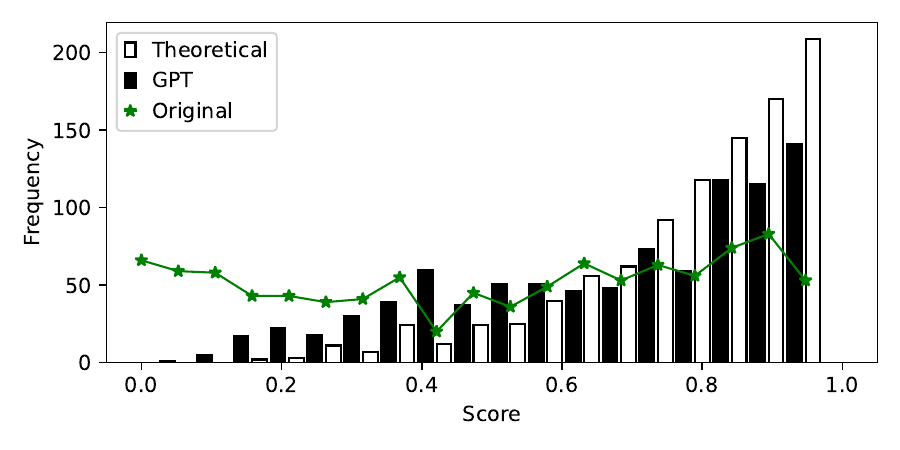}
        \caption{GPT-4.1}
        \label{subfig:q_hiring_d2_41}
    \end{subfigure}
    \hfill
    \begin{subfigure}{0.32\textwidth}
        \centering
        \includegraphics[width= \linewidth]{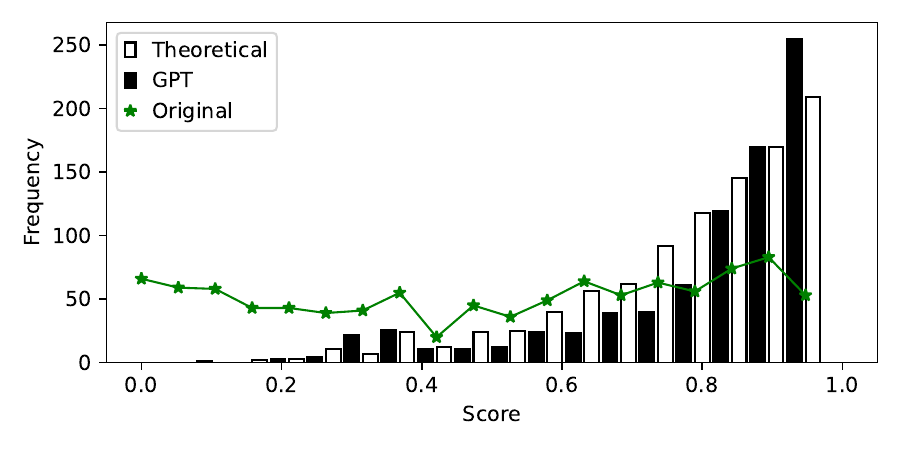}
        \caption{GPT-5}
        \label{subfig:q_hiring_d2_5}
    \end{subfigure}
    \caption{Comparison of the qualification distribution before/after agent response in the law school setting and decision scenario 2.}
    \label{fig:q_hiring_d2}
\end{figure*}

\begin{figure*}[h]
    \centering
    \begin{subfigure}{0.32\textwidth}
        \centering
        \includegraphics[width= \linewidth]{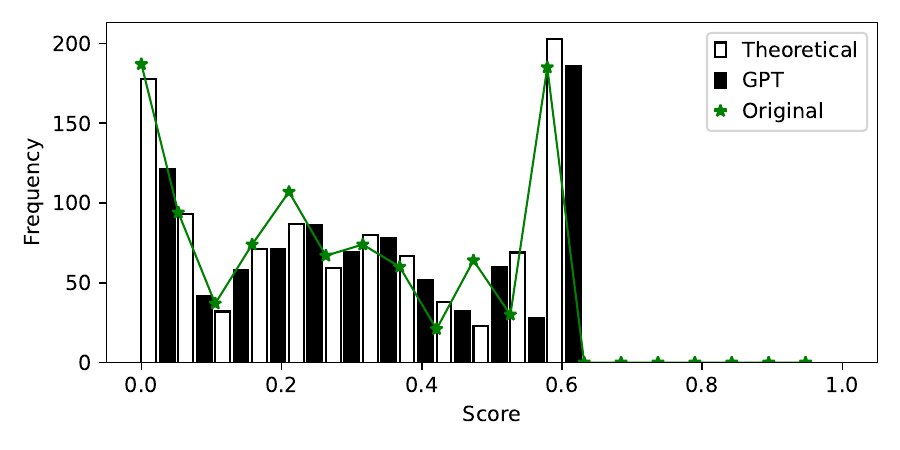}
        \caption{GPT-4o}
        \label{subfig:q_income_d2_4o}
    \end{subfigure}
    \hfill
    \begin{subfigure}{0.32\textwidth}
        \centering
        \includegraphics[width= \linewidth]{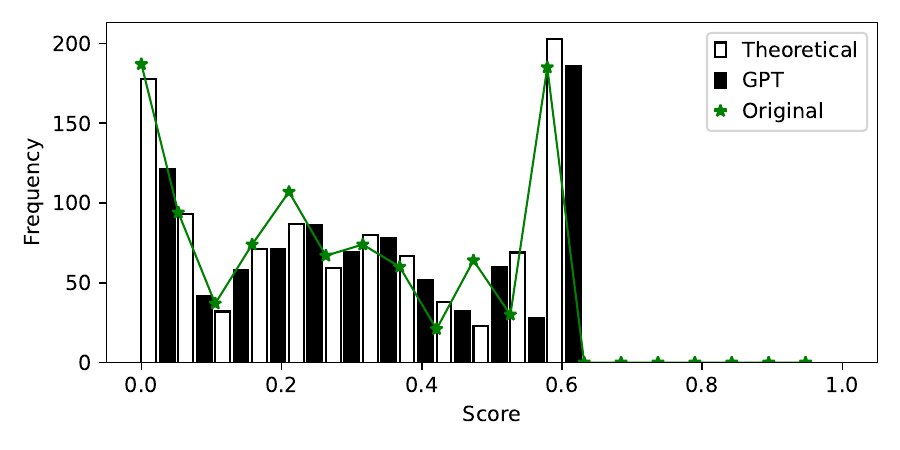}
        \caption{GPT-4.1}
        \label{subfig:q_income_d2_41}
    \end{subfigure}
    \hfill
    \begin{subfigure}{0.32\textwidth}
        \centering
        \includegraphics[width= \linewidth]{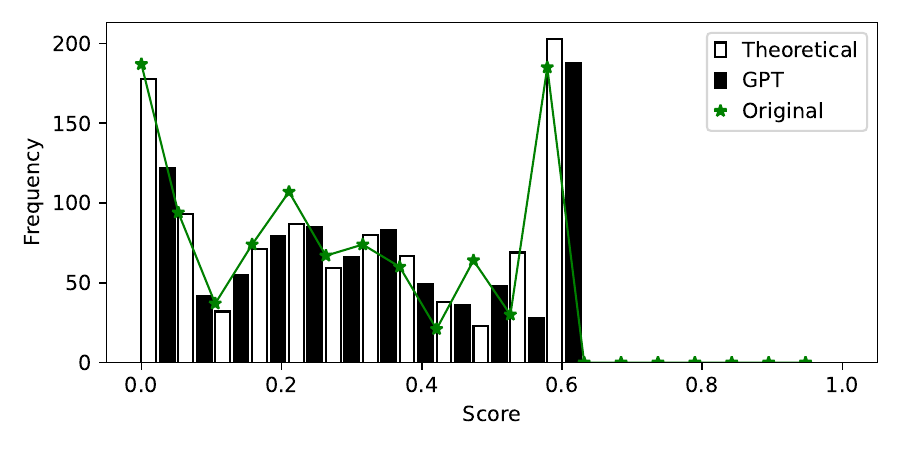}
        \caption{GPT-4.1}
        \label{subfig:q_income_d2_5}
    \end{subfigure}
    \caption{Comparison of the qualification distribution before/after agent response in the income setting and decision scenario 2.}
    \label{fig:q_income_d2}
\end{figure*}

\begin{figure*}[h]
    \centering
    \begin{subfigure}{0.32\textwidth}
        \centering
        \includegraphics[width= \linewidth]{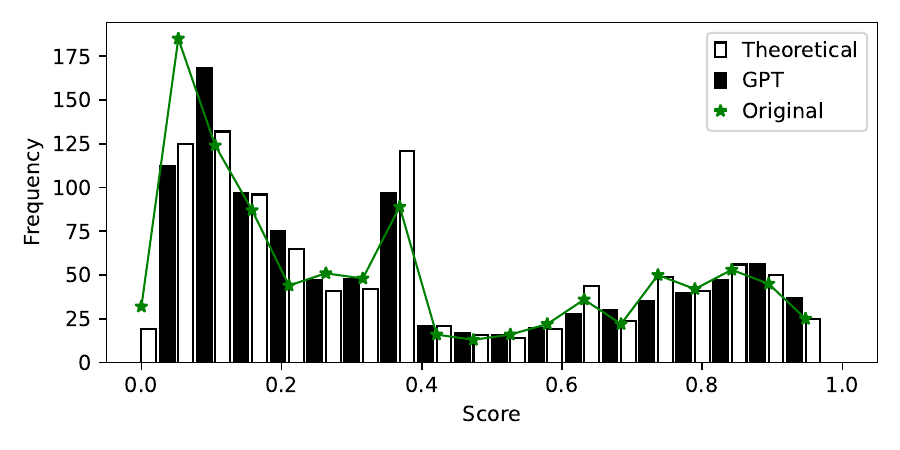}
        \caption{GPT-4o}
        \label{subfig:q_credit_d2_4o}
    \end{subfigure}
    \hfill
    \begin{subfigure}{0.32\textwidth}
        \centering
        \includegraphics[width= \linewidth]{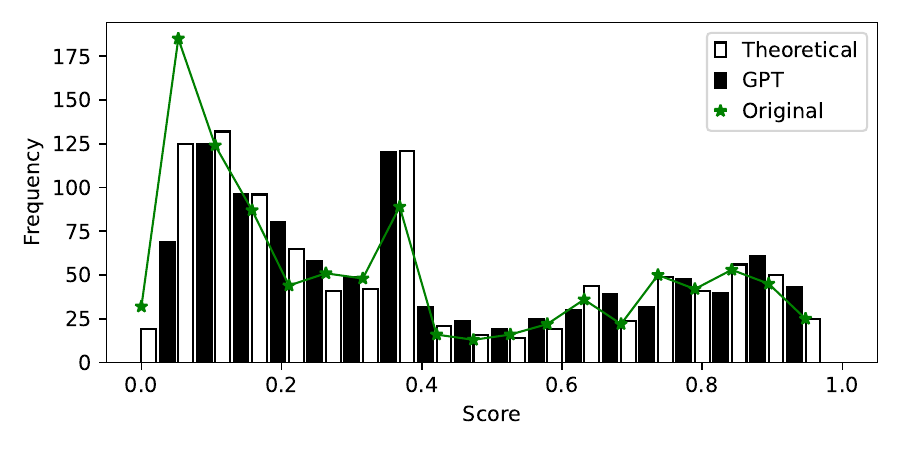}
        \caption{GPT-4.1}
        \label{subfig:q_credit_d2_41}
    \end{subfigure}
    \hfill
    \begin{subfigure}{0.32\textwidth}
        \centering
        \includegraphics[width= \linewidth]{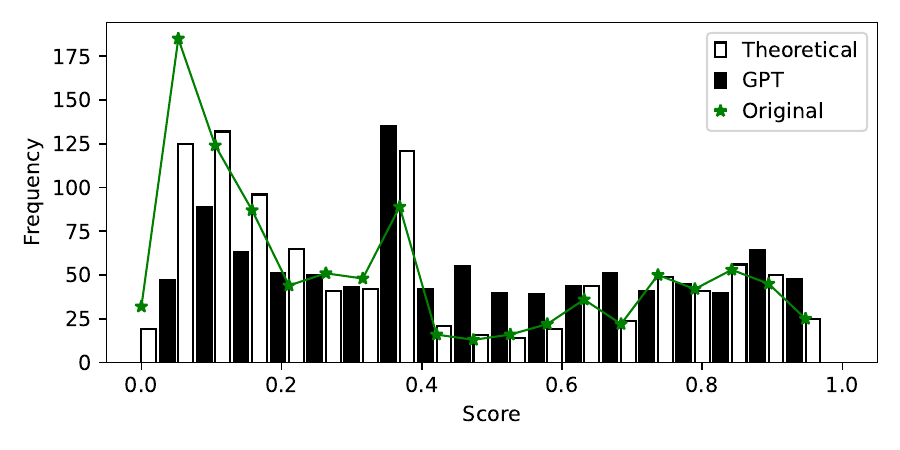}
        \caption{GPT-5}
        \label{subfig:q_credit_d2_5}
    \end{subfigure}
    \caption{Comparison of the qualification distribution before/after agent response in the loan approval setting and decision scenario 2.}
    \label{fig:q_credit_d2}
\end{figure*}

\begin{figure*}[h]
    \centering
    \begin{subfigure}{0.32\textwidth}
        \centering
        \includegraphics[width= \linewidth]{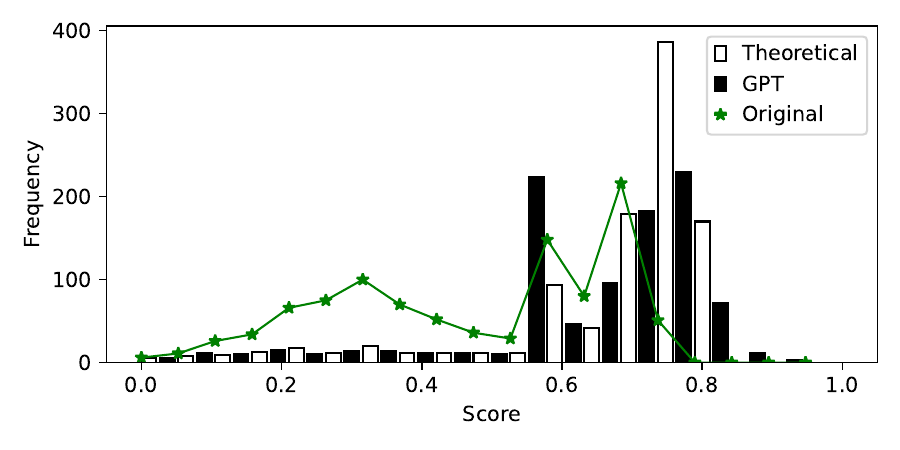}
        \caption{GPT-4o}
        \label{subfig:q_public_d2_4o}
    \end{subfigure}
    \hfill
    \begin{subfigure}{0.32\textwidth}
        \centering
        \includegraphics[width= \linewidth]{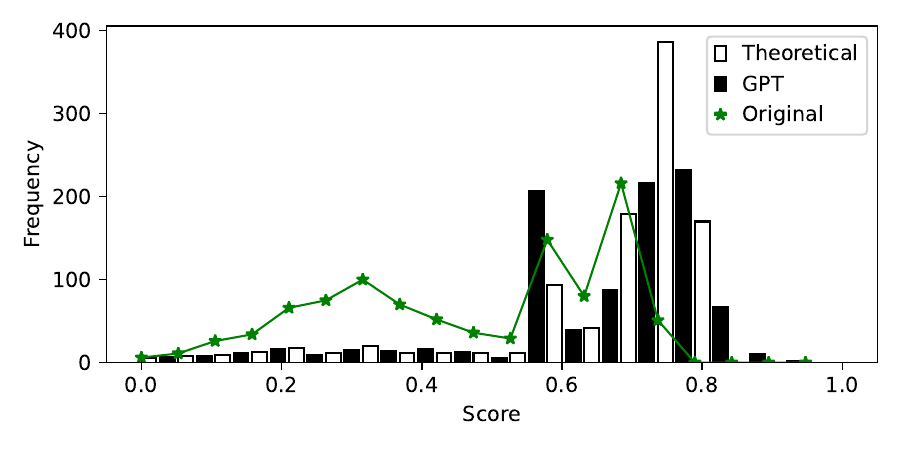}
        \caption{GPT-4.1}
        \label{subfig:q_public_d2_41}
    \end{subfigure}
    \hfill
    \begin{subfigure}{0.32\textwidth}
        \centering
        \includegraphics[width= \linewidth]{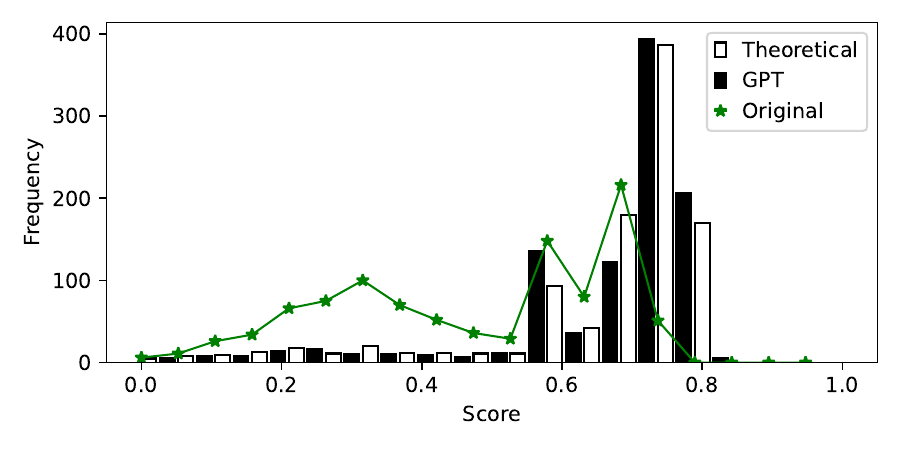}
        \caption{GPT-5}
        \label{subfig:q_public_d2_5}
    \end{subfigure}
    \caption{Comparison of the qualification distribution before/after agent response in the public assistance program setting and decision scenario 2.}
    \label{fig:q_public_d2}
\end{figure*}

Plots of score distribution changes are shown in Fig. \ref{fig:score_income_d1} to Fig. \ref{fig:score_pap_d1}, while plots of qualification distribution changes are shown in Fig. \ref{fig:q_credit_d1} to Fig. \ref{fig:q_public_d2}.

\begin{figure*}[h]
    \centering
    \begin{subfigure}{0.32\textwidth}
        \centering
        \includegraphics[width= \linewidth]{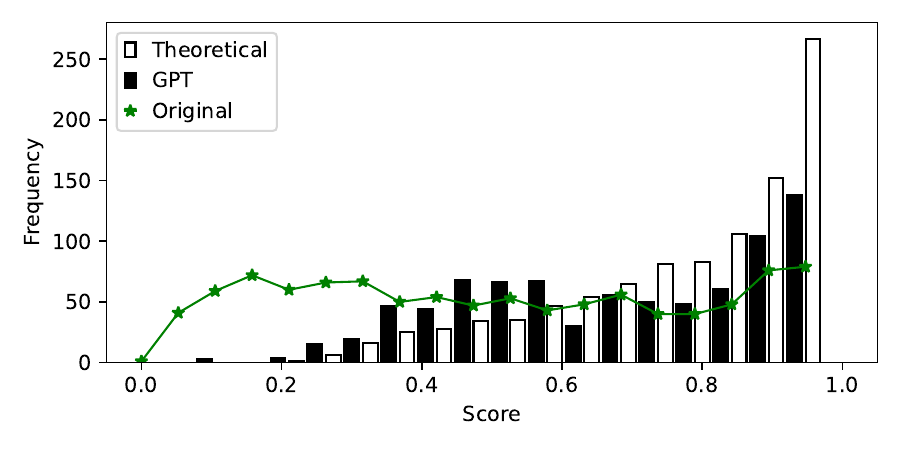} 
        \caption{GPT-4o}
        \label{subfig:score_hiring_d1_4o}
    \end{subfigure}
    \hfill
    \begin{subfigure}{0.32\textwidth}
        \centering
        \includegraphics[width= \linewidth]{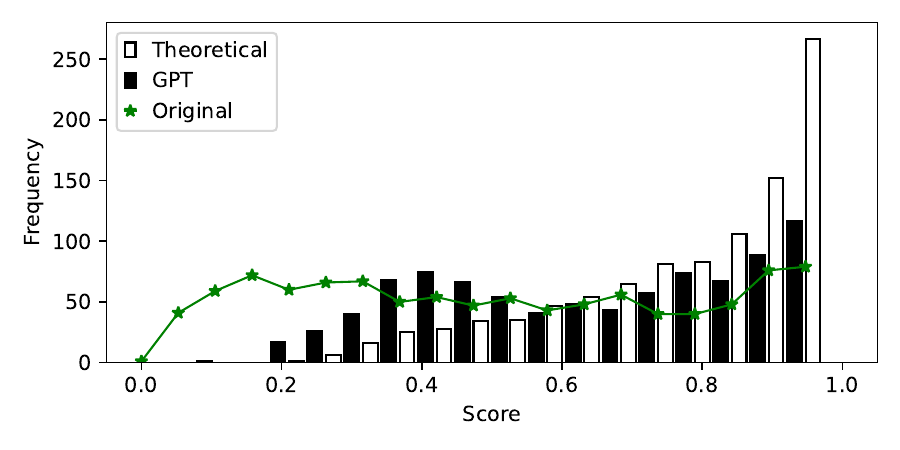}
        \caption{GPT-4.1}
        \label{subfig:score_hiring_d1_41}
    \end{subfigure}
    \hfill
    \begin{subfigure}{0.32\textwidth}
        \centering
        \includegraphics[width= \linewidth]{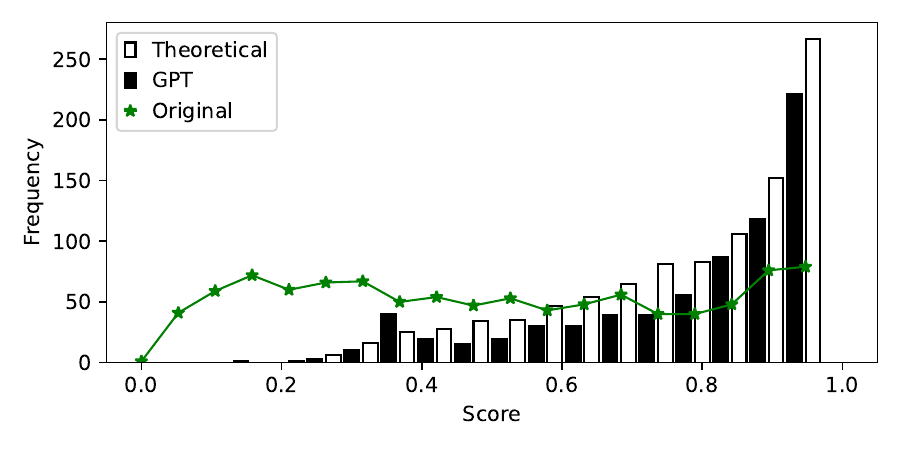}
        \caption{GPT-5}
        \label{subfig:score_hiring_d1_5}
    \end{subfigure}
    \caption{Comparison of the score distribution before/after agent response in the law school setting.}
    \label{fig:score_hiring_d1}
\end{figure*}

\begin{figure*}[h]
    \centering
    \begin{subfigure}{0.32\textwidth}
        \centering
        \includegraphics[width= \linewidth]{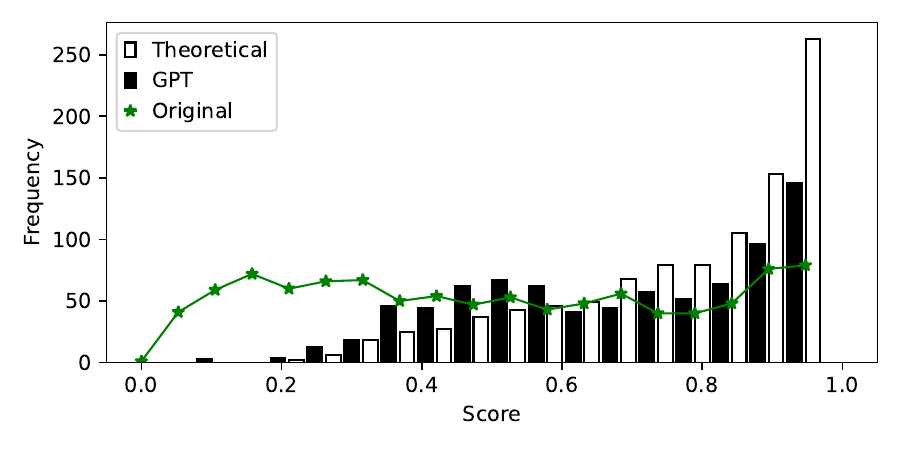}
        \caption{GPT-4o}
        \label{subfig:q_hiring_d1_4o}
    \end{subfigure}
    \hfill
    \begin{subfigure}{0.32\textwidth}
        \centering
        \includegraphics[width= \linewidth]{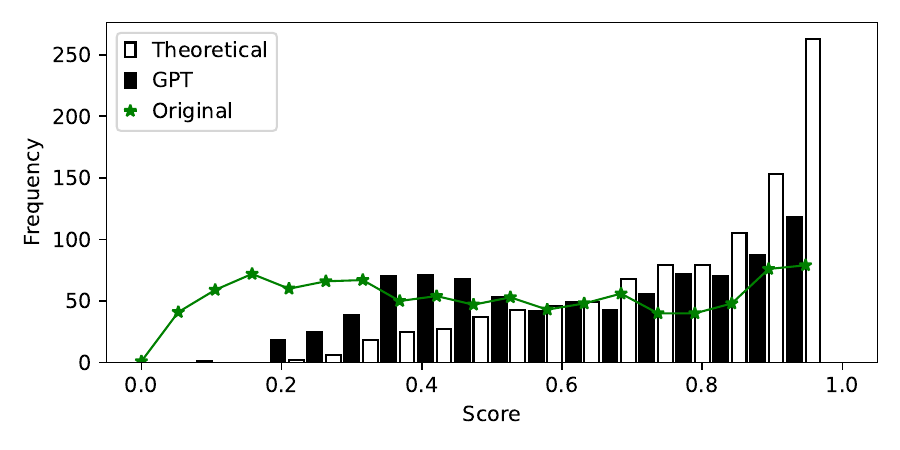}
        \caption{GPT-4.1}
        \label{subfig:q_hiring_d1_41}
    \end{subfigure}
    \hfill
    \begin{subfigure}{0.32\textwidth}
        \centering
        \includegraphics[width= \linewidth]{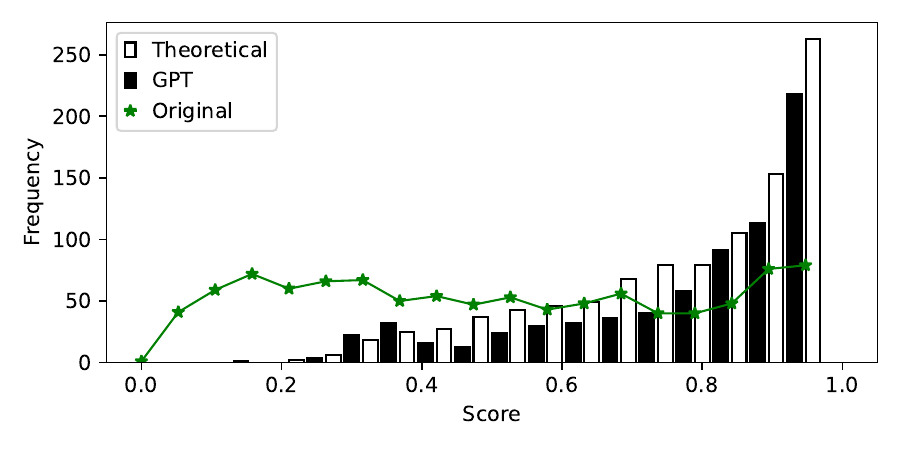}
        \caption{GPT-5}
        \label{subfig:q_hiring_d1_5}
    \end{subfigure}
    \caption{Comparison of the qualification distribution before/after agent response in the hiring setting and for decision scenario 1.}
    \label{fig:q_hiring_d1}
\end{figure*}

\begin{figure*}[h]
    \centering
    \begin{subfigure}{0.32\textwidth}
        \centering
        \includegraphics[width= \linewidth]{Plots/4oincome_strategy_score_dist.pdf}
        \caption{GPT-4o}
        \label{subfig:q_income_d1_4o}
    \end{subfigure}
    \hfill
    \begin{subfigure}{0.32\textwidth}
        \centering
        \includegraphics[width= \linewidth]{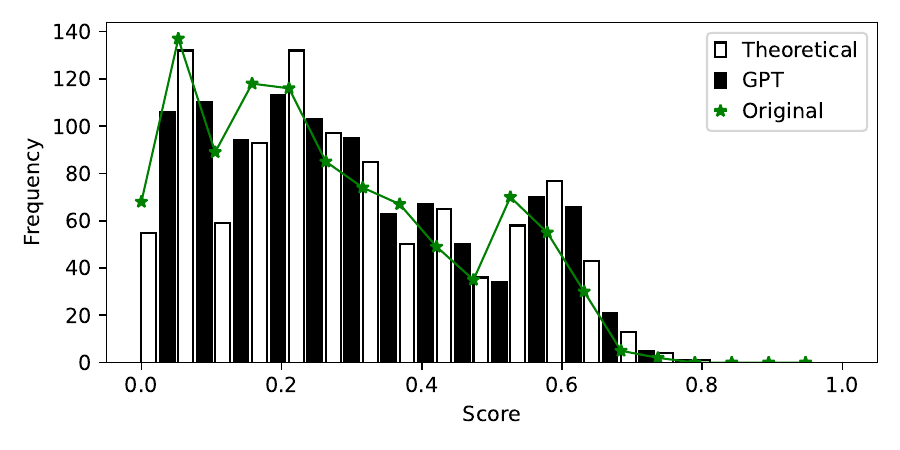}
        \caption{GPT-4.1}
        \label{subfig:q_income_d1_41}
    \end{subfigure}
    \hfill
    \begin{subfigure}{0.32\textwidth}
        \centering
        \includegraphics[width= \linewidth]{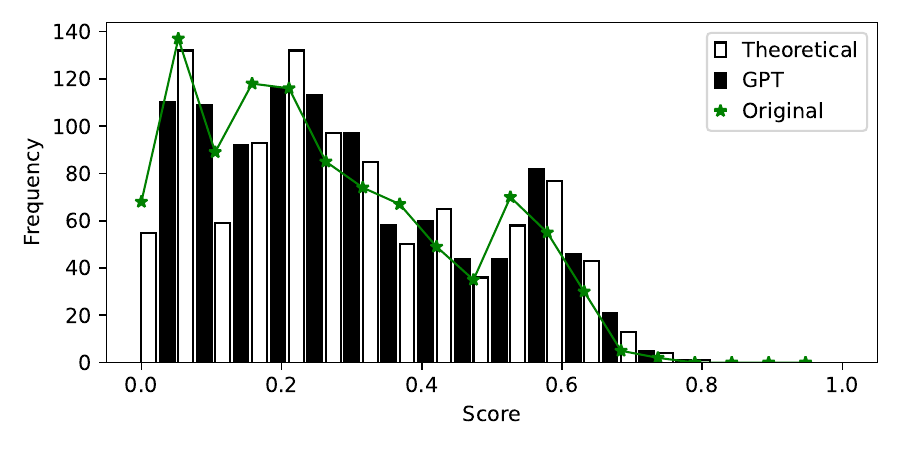}
        \caption{GPT-5}
        \label{subfig:q_income_d1_5}
    \end{subfigure}
    \caption{Comparison of the qualification distribution before/after agent response in the hiring setting and for decision scenario 1.}
    \label{fig:q_income_d1}
\end{figure*}

\end{document}